# A comprehensive review of classifier probability calibration metrics

Richard Oliver Lane

QinetiQ, St. Andrews Road, Great Malvern, Worcestershire, UK

**Abstract**: Probabilities or confidence values produced by artificial intelligence (AI) and machine learning (ML) models often do not reflect their true accuracy, with some models being under or over confident in their predictions. For example, if a model is 80% sure of an outcome, is it correct 80% of the time? Probability calibration metrics measure the discrepancy between confidence and accuracy, providing an independent assessment of model calibration performance that complements traditional accuracy metrics. Understanding calibration is important when the outputs of multiple systems are combined, for assurance in safety or business-critical contexts, and for building user trust in models. This paper provides a comprehensive review of probability calibration metrics for classifier and object detection models, organising them according to a number of different categorisations to highlight their relationships. We identify 82 major metrics, which can be grouped into four classifier families (point-based, bin-based, kernel or curve-based, and cumulative) and an object detection family. For each metric, we provide equations where available, facilitating implementation and comparison by future researchers.

**Keywords**: Classification, Calibration, Confidence, Object Detection, Multiclass, Uncertainty.

## 1 Introduction

Artificial intelligence (AI) and machine learning (ML) models have seen widespread adoption in recent years. When such models are used in safety or business-critical applications, it is vital to be able to understand and assure their behaviour. Models generate predictions accompanied by confidence scores or probabilities, which are considered calibrated when they accurately reflect the proportion of correct classification decisions. However, confidence scores are not always representative of true probabilities. Assessing model calibration, particularly under operational conditions differing from training, requires robust measures of calibration quality.

The publication of [24], which highlighted examples of miscalibration in deep neural networks, sparked an intense interest in the concept of calibration and metrics to measure it. This surge in attention has led to a large number of publications over the last few years, with approximately ten new metrics being defined and proposed every year since then, and many papers discussing their merits or application. However, as Flach [16] witheringly puts it "contrary what recent machine learning literature may lead you to believe, calibration research predates machine learning and has been studied for three-quarters of a century". Thus, there is a large body of work from which to draw on. A recent survey on assessing and improving classifier calibration was conducted in [69]. The authors were motivated to write the survey as "the literature on post-hoc classifier calibration in machine learning is now sufficiently rich that it is no longer straightforward to obtain or maintain a good overview of the area". This lack of understanding has caused a bottleneck, as researchers have to search many individual papers to find relevant information. Indeed, despite the wide-ranging and valuable survey of [69], it only covers a subset of the metrics available at the time. A survey on more general metrics (not just calibration) for image analysis by a large group of experts includes a flow chart for deciding which calibration metric to use in different scenarios [46]. The paper describes the field of calibration metric development as "relatively young and currently highly dynamic". As such, the paper only covers a small number of better-known metrics. A comparison of variants of ten calibration metrics is given in [72], computing the correlation between the metrics for a number of classification tasks. Again, this only covers the most well-known metrics.

This paper gives a comprehensive review of probability calibration metrics for classifier and object detection models with a discrete sets of outputs, organising the metrics according to a number of different categorisations to help understand the relationships between them. We have identified 82 major metrics, which can be grouped into four classifier families (point-based, bin-based, kernel or fitted-curve, and cumulative) and an object detection family. Each family has advantages and disadvantages, as do the individual metrics within the families. This paper represents the most comprehensive survey of probability calibration metrics to-date, providing descriptions of significantly more metrics than [16], [46], and [69] combined. Where available, equations are provided to facilitate implementation and comparison by future researchers. This categorisation and definition of metrics will serve as a useful reference for the high volume of researchers currently attempting to understand the relationships and differences between different metrics. The survey concentrates on the assessment of model probabilities rather than methods to improve calibration, which is covered well by [16] and [69].

The remainder of this paper is organised as follows. Section 2 introduces key concepts relating to calibration metrics and defines the notation used in this paper. Sections 3 to 6 respectively describe metrics in the point-based, bin-based, kernel or fitted-curve, and cumulative families. Section 7 discusses metrics designed specifically to assess object detection algorithms. A dendrogram showing the hierarchy of families and sub-families, as described in sections 3 to 7, is shown in Fig. 1. Section 8 describes techniques for visualising probability calibration. Section 9 discusses general frameworks and theories for understanding calibration metrics. Finally, conclusions are given in section 10. Appendix A contains a table that summarises the main pros and cons of each metric and other information, such as the range of attainable values and alternative names for the same metric.

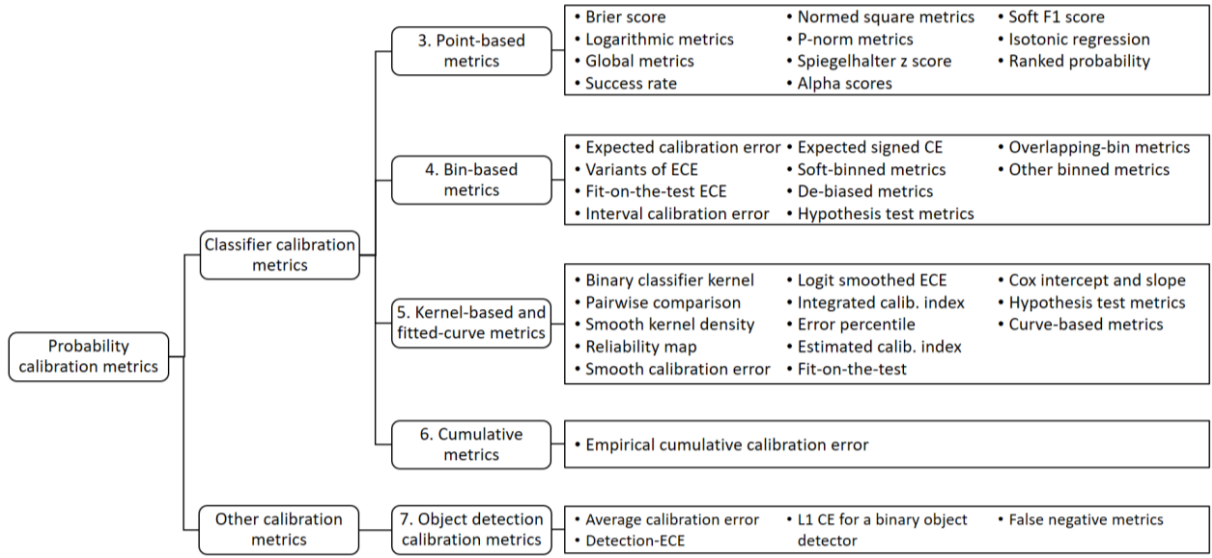

**Fig. 1** Hierarchy of probability calibration metric families and sub-families, as described in sections 3 to 7.

## 2 Calibration metric concepts

### 2.1 Notation

The literature on probability calibration metrics is somewhat inconsistent in its terminology and notation. Where possible, we describe the techniques in a unified way in this paper, rather than using the original symbolization. We start by defining the problem of interest to be assigning a piece of data or "data point" to one of $K$ classes. The data could be a vector of features, an image, a signal, a time series, or any other collection of information. Each data point, while potentially representing multiple numbers, is treated here as a single, possibly multi-dimensional, entity. Each data point $i$ has an associated true label $y_i$. If the classifier is binary, then $y \in \{0,1\}$ by convention. If the classifier is multi-class (with three or more classes), then $y \in \{1, \ldots, K\}$ or is encoded as a one-hot vector, as determined by context. When multi-class problems are decomposed into multiple binary problems, $y_i$ represents making a correct decision for a particular class. The number of data points over which the calibration of a model is to be assessed is $N$. The vector $\bar{y}$ is a length-$K$ vector of true proportions of each class in the dataset.

For each data point, the model produces a predicted probability or confidence $c_i$. For binary problems, by convention this is a scalar value representing the probability of class 1. In this case, there is no need to specify the probability of class 0 as it is completely determined by $1 - c_i$. This simplifies the analysis. For multi-class problems, $c_i$ should be considered to be a vector of probabilities for each class, which sums to unity. When discussing the extension of binary classification to multi-class classification, the multi-class notation may be used for binary classes. That is, $c_i$ may be a two-element vector representing the probabilities of the two classes.

A list of symbols used in this paper and their basic definition is given in Table 1. Each term is described in more detail where it first appears in the paper.

| Symbol | Definition |
| --- | --- |
| $a_b$ | Soft-binned accuracy |
| $B$ | Number of bins |
| $b$ | Bin index |
| $c_i$ | Confidence of data point $i$ |
| $\bar{c}_b$ | Mean confidence in bin $b$ |
| $\tilde{c}(c)$ | Kernel density estimate of confidence |
| $G$ | Number of grid points |
| $g$ | Grid point index |
| $H$ | Number of proximity bins |
| $h$ | Kernel bandwidth or proximity bin index, according to context |
| $h(c)$ | Logit function |
| $h_{ij}$ | Pairwise error term |
| $I[\cdot]$ | Indicator function |
| $i$ | Data point index |
| $j$ | Second class index or data point index, according to context |
| $K$ | Number of classes |
| $k$ | Class index or other index, according to context |
| $k(\cdot)$ | Kernel function |
| $k^*$ | Max parameter in surrogate interval calibration error |
| $L(\cdot)$ | Loss function |
| $M$ | Number of Monte Carlo runs |
| $m$ | Reliability map cluster size or Monte Carlo run index |
| $N$ | Total number of data points |
| $n_b$ | Number of data points in bin $b$ |
| $p$ | Parameter of the p-norm |
| $p_b$ | Proportion of data points in bin $b$ |
| $Q$ | Number of partitions of data |
| $q$ | Partition index |
| $R_{bm}$ | Random sample from label distribution |
| $R(\bar{c})$ | Reliability map estimate |
| $s$ | Number of data points in a sliding bin |
| $(s)$ | Sampled value |
| $s_b$ | Soft-binned size |
| $s(c, \bar{y})$ | Scoring function |
| $T$ | Transpose operator |
| $u(c)$ | Bin membership function |
| $V$ | Number of detections |
| $v$ | Detection index |

| Symbol | Definition |
|---|---|
| $w$ | Bin width |
| $w_0$ | Biases in a model |
| $w_1$ | Weights in a model |
| $w_k$ | Class importance |
| $y_i$ | Class label of data point $i$ |
| $x$ | Input data |
| $\bar{y}$ | True proportions of each class in the dataset |
| $\hat{y}$ | Calibration map |
| $\bar{y}_b$ | Mean accuracy in bin $b$ |
| $Z$ | Normalisation constant |
| $\alpha$ | Exponent in alpha scores or other exponent, according to context |
| $\gamma$ | Focus parameter |
| $\delta_{i,j}$ | Kronecker delta function |
| $\varepsilon$ | Small value |
| $\Theta$ | Set of eligible bins |
| $\mu$ | Mean |
| $\rho(\cdot)$ | Inverse logit function |
| $\sigma$ | Standard deviation |
| $\tau$ | Temperature parameter |
| # | Number of elements (cardinality) of a set |

**Table 1** Notation for probability calibration used in this paper.

## 2.2 Calibration curve and reliability diagram

An ideally calibrated classifier outputs confidence scores or predicted probabilities equal to its accuracy. However, in practice the accuracy for a particular confidence value can be higher or lower than that. The actual accuracy as a function of confidence for a particular class is known as the calibration curve. An example theoretical non-perfect calibration curve is illustrated in Fig. 2, along with the ideal calibration line. In this case, the classifier is over-confident in the target class – the achieved accuracy is lower than the model's confidence. In binary classification, the calibration curve for class 1 contains all the information necessary to understand the model's calibration, as the curve for class 0 is its mirror image reflected over the perfect calibration line. For multi-class problems, the situation is more complex; this is explained further below.

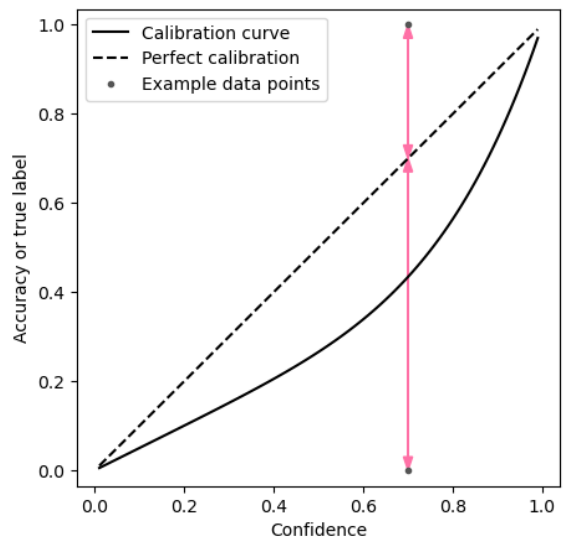

**Fig. 2** Theoretical calibration curve with two example data points having the same confidence, but different true labels.

For a single data point, the classifier is either correct or incorrect and the model can only perfectly be calibrated if the confidence is zero or unity. For other confidence values, there is inevitably some calibration error. Fig. 2 demonstrates this for two example data points with the same confidence value of 0.7 for class 1. The true label is zero for one data point and the error for that data point is 0.7. The true label for the other data point is unity, and the error for that data point is 0.3.

Although perfect calibration for a single data point is impossible without perfect accuracy, a classifier with non-perfect accuracy over a set of data points may still be well calibrated. For example, if ten data points all have a confidence of 0.7 and seven out of those ten are classified correctly, then this classifier is perfectly calibrated for the dataset. In practice, not all confidence values output by a model are expected to be identical. It is therefore common to group confidence values into non-overlapping bins and analyse the accuracy of data points in each bin. When this information is presented visually it is known as a reliability diagram, curve, or plot. This is frequently shown as a bar chart, see [24] for example. However, line plots facilitate better visual inspection of the data. Some works place markers at the centre of each bin on the horizontal confidence axis. However, if the data is not uniformly distributed throughout the bin this can give misleading results. Therefore, it is better to place markers at the mean confidence for each bin [69]. Vasilev et al. [77] distinguish between the bar chart representation being called the reliability diagram and the line representation being called the reliability plot.

An example reliability plot is shown in Fig. 3 as a solid line, and additional information is included. The dataset used to compute this diagram was generated as 500 random samples, with the confidence and true labels determined based on the true calibration curve in Fig. 2. In Fig. 3, true labels are jittered by ±0.025 for improved visualization. The markers on the reliability curve show the mean achieved accuracy in ten equal-width bins, and the error bars represent the standard error of those estimates. The standard error is greater for confidence values near 0.5 than zero or unity, and is in general greater when there are fewer data points, although this effect is not apparent for this example dataset. The measured reliability curve has approximately the same shape as the true calibration curve. The same information is shown in Fig. 4 using the slightly more common bar plot representation.

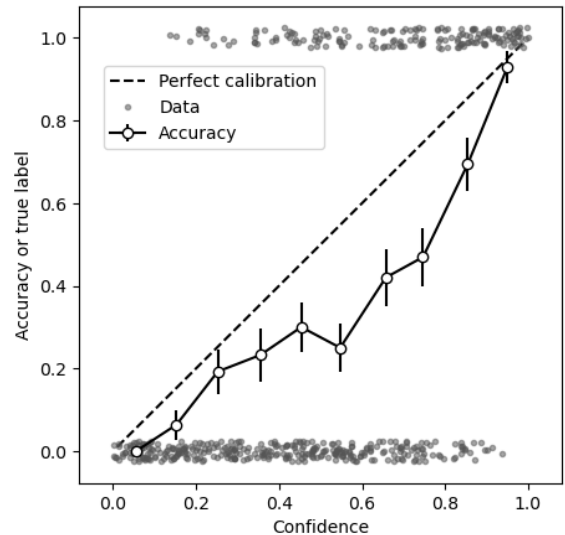

**Fig. 3** Reliability line plot for 500 labelled data points, with ten bins.

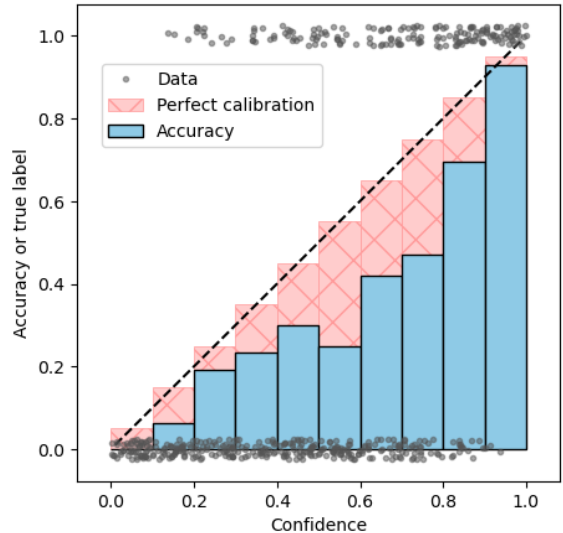

**Fig. 4** Reliability bar chart for 500 labelled data points, with ten bins.

A grouping diagram is a modified version of the reliability plot [61]. In addition to the mean accuracy for each whole bin, the diagram shows the mean accuracy for two different groups of data points in each bin. The groups are partitions of the input feature space, chosen to maximise the difference in their accuracy. This allows the visualisation of data from different groups whose individual over- and under-confidences may cancel out to give the appearance of a well calibrated system. However, unlike the standard reliability plot, this uncommon diagram requires access to data feature values, which may not be output by some models.

Most binary calibration metrics measure some aspect of the data visible in Fig. 2, Fig. 3, or Fig. 4, whether this is the location of the data points, or the degree to which the calibration curve (as estimated via binning or other means) deviates from the identity line.

## 2.3 Multi-class issues

Most work on probability calibration metrics relates to binary classifiers. However, multi-class versions also exist. There are three ways to define calibration for multiple classes [69]:

- **Confidence**. Confidence calibration considers only the class with the highest predicted probability. Some classifiers only report probabilities for the most likely class. In those situations this is the only calibration that can be assessed. This is also known as top-label calibration, and is the most common form of calibration.
- **Class-wise**. Class-wise calibration assesses the marginal probabilities of all classes. This definition requires all one-vs-rest binary estimators to be calibrated individually. This is a more restrictive definition than confidence as it is not possible for under and over estimates of probability from different classes to cancel out.
- **Multiclass**. Multiclass calibration requires the entire vector of probabilities to be correct simultaneously. This is also known as joint, full, canonical, or strong calibration.

Vaicenavicius et al. [75] give theoretical examples where a multiclass classifier is either confidence-calibrated or class-wise calibrated, but not fully multiclass calibrated. Lack of full multiclass calibration could be important in safety critical applications, especially where the action taken should depend not only on the most likely outcome, but the probabilities of other less-likely outcomes, which may have severe negative consequences.

Multi-class problems are often decomposed into a set of binary sub-problems whose outputs are aggregated to give an overall multi-class score [35]. The most common decompositions are one-vs-rest (OVR) or comprehensive pairwise. The decomposition can be represented in code matrices, whose columns represent sub-problems and rows represent classes. Entries are +1 for the positive class, -1 for the negative class, and 0 if the class is not represented in the sub-problem. Example code matrices for one-vs-rest and pairwise decompositions for a four-class problem are respectively shown in (1) and (2).

$$M_{ovr} = \begin{bmatrix} +1 & -1 & -1 & -1 \\ -1 & +1 & -1 & -1 \\ -1 & -1 & +1 & -1 \\ -1 & -1 & -1 & +1 \end{bmatrix} \quad (1)$$

$$M_{pair} = \begin{bmatrix} +1 & +1 & +1 & 0 & 0 & 0 \\ -1 & 0 & 0 & +1 & +1 & 0 \\ 0 & -1 & 0 & -1 & 0 & +1 \\ 0 & 0 & -1 & 0 & -1 & -1 \end{bmatrix} \quad (2)$$

Other code matrix aggregation setups are possible but are far less common.

## 2.4 Hypothesis tests

According to the metric of choice, the question may be asked: is this classifier calibrated? Most metrics are theoretically zero for a perfectly calibrated system, but due to randomness in the data and using finite datasets, the measured value of a metric may be nonzero. The question then becomes whether a non-zero value is statistically different from zero, under the assumption that the model is perfectly calibrated – the null hypothesis. The p-value of the metric is the probability under the null hypothesis that values at least as extreme as the measured value would occur in an experiment. The p-value can be used to perform a statistical significance test or hypothesis test: if it is less than a certain threshold (usually 0.05) the null hypothesis is rejected. However, modern practice in this situation is not to reject the null hypothesis outright, but to use the p-value to understand the data [82]. Indeed, with a big enough sample size, the null hypothesis will inevitably be rejected, as real datasets do not match theoretical distributions perfectly [27].

Hypothesis tests can also compare the calibration performance of two models. Such tests aim to evaluate whether the observed differences in calibration metrics are statistically significant. However, due to difficulties in calculating a suitable approximation to the null hypothesis, this is much less common. A number of metrics have a statistical hypothesis associated with them. Where that is the case, this is indicated in the appropriate section below.

## 2.5 Bootstrapping and consistency sampling

All of the metrics in this review have some inherent randomness. While the calculation of the metric is usually deterministic given a fixed dataset, the dataset itself is random. Furthermore, some metrics, such as the surrogate interval calibration error, are random, even given a fixed dataset. Therefore, it is of importance to know the uncertainty in the metric to determine the significance of different values.

Some metrics have known distributions, or at least good approximate known distributions. In those cases the p-values can be computed directly. Where the distribution is unknown, Monte Carlo methods can be used to estimate the uncertainty. Two methods for doing this are bootstrapping and consistency resampling [75].

Bootstrapping creates a number of new datasets $\{y_i^{(s)}, c_i^{(s)}\}$ from the original one $\{y_i, c_i\}$ by sampling with replacement. The metric is computed for each bootstrapped dataset and the variability in computed values is used to understand the uncertainty in this metric.

Consistency resampling, a modified version of bootstrapping, assumes that there is a known calibration map $\hat{y}(c)$, which in practice can be estimated from the data. Bootstrapping is used to generate samples just for $c$. The calibration map is used for each bootstrapped sample $c_i^{(s)}$ to produce $\hat{y}\left(c_i^{(s)}\right)$. This is then used to randomly select a label $y_i^{(s)}$ with probability $\hat{y}$. The final bootstrapped dataset is composed of pairs of values $\{y_i^{(s)}, c_i^{(s)}\}$. In contrast to standard bootstrapping, consistency resampling produces labels that were not necessarily in the original dataset but are consistent with the calibration map. This method allows smaller datasets to be used than standard bootstrapping, at the expense of requiring the calibration map to be estimated.

## 2.6 Other observations on calibration

Numerous studies have empirically shown that classifiers often exhibit overconfidence. The usual explanation given is that the models are large enough to memorize training data and maximise the confidence. However, Bai et al. [4] show that certain classifiers are inherently overconfident, even when the dataset is large and the number of model parameters is small. Specifically, this applies to logistic regression and other classifiers where the activation function is symmetric and concave in the positive half. Under-confident classifiers can arise when the activation function is convex in the positive half, but convexity over this whole range is not possible, so under-confidence cannot happen at every value of $c > 0.5$. Munir et al. [52] state that rectified linear unit (ReLU) activation functions, widely used in modern deep neural nets, and similar piece-wise linear functions, are a core reason behind overconfident predictions far away from the training data. However, Minderer et al. [49] show that model structure is more important than model size in understanding probability calibration, and that recent models, especially those not using convolutions, are actually among the best calibrated.

A common type of "probabilistic" classifier is the Naïve Bayes classifier as it naturally outputs ostensible probabilities. However, as pointed out by Flach [15], due to its assumptions, it is not well calibrated and is technically neither probabilistic nor Bayesian.

The metrics described in this paper serve as absolute measures of model calibration. Recalibration techniques attempt to improve the correctness of model confidence values by post processing those values. The ideal aim of such techniques is to invert the effect of the model calibration curve so that the overall process produces an identity calibration curve. In practice, these techniques are not perfect, but it is possible to define their "calibration gain", which is the improvement in calibration error (by any particular metric) when the technique is applied [88].

## 2.7 Families of metrics

This survey groups classifier confidence metrics into four families: point-based, bin-based, kernel or curve-based, and cumulative. Point-based metrics compute a score for each data point $(y_i, c_i)$ and aggregate these scores to give an overall calibration measure for the whole dataset. Although easy to define, point-based metrics can only achieve perfect calibration scores when a classifier is 100% confident and is always correct. For lower confidence scores there is always some discrepancy between the confidence and the correct decision for a particular data point, as illustrated by the data point calibration error in Fig. 2. Point-based metrics are described in more detail in section 3.

Bin-based metrics group data-points with similar confidence values into bins and compare the proportion of correct classification in a bin to the mean confidence of the bin. This allows non-perfect, but well calibrated, classifiers to achieve a low calibration score. An illustration of binned data is shown in Fig. 4. Conceptually, bin-based metrics are based on the difference between the binned accuracy and perfect calibration lines. A disadvantage of bin-based metrics is that a small change in confidence value of one data point can cause it to move to an adjacent bin, resulting in a discrete jump in metric value, which may be undesirable. Bin-based metrics are described in more detail in section 4.

Kernel or curve-based metrics fit a smooth calibration curve to the data and compare this to the perfect calibration line. In Fig. 2, the calibration error is based on the area between the fitted calibration curve and the perfect line. With kernel or curve-based metrics, small changes in confidence values of individual data points result in small changes in metric values. Different metrics within this family use different models of varying complexity to fit the data. Kernel and curve-based metrics are described in more detail in section 5.

Cumulative metrics sort data points by confidence and examine the difference between the cumulative accuracy and the perfect calibration line. The advantage of such metrics is that there is no need to set or estimate arbitrary parameters, such as bin width or parameters of a curve model. Cumulative metrics are described in more detail in section 6.

Beyond classifiers, a common type of probabilistic model is an object detector. Whereas as a classifier only considers the discrete label of a piece of data, an object detector also provides object size and location co-ordinates, often within an image. Although it is possible to assess object detector confidence scores based only on labels of associated ground-truth objects, the additional degrees of freedom allow a more nuanced form of assessment. Object detection calibration metrics are described in more detail in section 7.

# 3 Point-based metrics

## 3.1 Introduction to point-based metrics

Point-based metrics compute a score for each data point $(y_i, c_i)$ and aggregate these scores to give an overall calibration measure for the whole dataset. Point-based metrics are usually simple to compute and are thus often more interpretable than some other types of metric. However, some of the individual score functions lack interpretability, despite having potentially useful mathematical properties, which may limit their use. Section 3.2 below describes the abstract concept of "proper scores", which provides a framework for analysing the properties of point-based metrics. The remainder of the section from 3.3 onwards describes specific point-based metrics.

## 3.2 Proper scores

Proper scoring rules are point-based evaluation measures for probability estimates that avoid the need for putting confidence scores into bins [69]. Following Bröcker [10], a scoring rule $S(c, y)$ is a function of a probability prediction (or confidence) $c$ and an outcome $y$. Let $\bar{y}$ be a length-$K$ vector representing the proportion in the data of each class. A scoring function, or simply score, is defined as:

$$s(c, \bar{y}) = \sum_{k=1}^{K} S(c, k) \bar{y}_k \qquad (3)$$

It is convention that low scores indicate good predictions. A score is "proper" if the divergence $s(c, \bar{y}) - s(\bar{y}, \bar{y})$ is nonnegative, and "strictly proper" if zero divergence implies $c = \bar{y}$. Informally, proper scores are optimal when predicted probabilities match the true proportions in the data [69].

Strictly proper scores can be decomposed into three terms: reliability (REL), resolution (RES), and uncertainty (UNC), which facilitate interpretation of the score [10]:

$$Score = REL - RES + UNC \qquad (4)$$

Reliability is a measure of the degree to which predictions differ from the actual sample relative frequencies [53], with lower values being good. In early works, reliability was referred to as validity [69]. However, it is often now known as calibration loss [69]. As an illustrative example, if a classifier only produces confidence values of exactly 90% or 70%, and is respectively correct 90% or 70% of the time for those specific values, then the classifier is reliable and has a reliability value of zero.

Resolution measures the degree to which sample proportions for each unique predicted probability differ from the overall sample proportions for the whole dataset [53], with higher values being better. For example, consider a binary dataset where 50% of the data points belong to either class. A classifier that always gives a confidence of 50% would be well-calibrated globally, but not particularly useful. This classifier has zero resolution. An alternative, reliable classifier that gives confidence scores of 80% or 20% half the time for either value has a better, non-zero resolution. Resolution is also known as sharpness [10].

Uncertainty is the score that would be achieved by replacing confidence values with the proportions of the actual samples [53], with lower being better. Thus, this term is inherent to the data and does not relate to the classifier. For example, if 80% of the data points are from one class, a theoretical classifier that always gives a confidence value of 80%, would achieve a score equal to the uncertainty value. If all data points are from one class, the uncertainty is zero. In a binary classifier, the uncertainty is highest when half of the data points belong to each class.

The three-term decomposition of scores is only useful when more than one data point has the same predicted probability vector. This may be the case for human forecasters that are prone to specifying probabilities on a discrete scale (10%, 20%, etc.). However, algorithms specify probabilities on a continuous scale from zero to unity and it is unlikely that many predicted probability vectors will have exactly the same value. When all probabilities are different, the resolution and uncertainty terms cancel out and the reliability is the same as the overall score. Thus the decomposition has less utility in modern algorithmic contexts than traditional human prediction analysis.

An alternative decomposition considers proper scores as a sum of two components: epistemic loss, due to the model not being optimal, and an irreducible or aleatoric loss, which is the loss of the theoretically optimal model, due to randomness in the data [69]. This decomposition helps focus analysis on parts of the problem that are in control of model designers.

Silva Filho et al. [69] recommend that classifiers should be trained using a proper scoring rule as loss function rather than non-proper functions. This is because the resulting models are likely to produce better probabilities, since probability refinement and calibration would be encouraged during training.

### 3.3 Brier score

The Brier score (BS) is one of the earliest and best-known calibration metrics. It is a common way of measuring how much the accuracy of a model diverges from its confidence, with lower scores meaning that the model is well calibrated. The Brier score is also known as the mean square error (MSE) [15] or the quadratic score [21]. The Brier score can be calculated by equation (5).

$$BS = \frac{1}{NK} \sum_{i=1}^{N} \sum_{k}^{K} (c_{ik} - y_{ik})^2 \qquad (5)$$

The score can be thought of as the mean of the squares of the arrowed data-point calibration-error line distances in Fig. 2.

Initially developed for weather forecasting, the Brier score is now widely used as a general measure of predictive performance and risk. It has advantages in that it is easily calculated and interpreted. The original definition of Brier score for binary classifiers is in the range [0,2]. Modern usage divides the classical definition by the number of classes to give a value always in the range [0,1] [69]. Wallace et al. [79] describe a stratified version of Brier score that captures class-specific calibration. Verhaeghe et al. [78] do not use the Brier score as it does not evaluate the "clinical value" of prediction models. They prefer to use expected calibration error (ECE) or its variants, which are described later in this paper.

The square root of the Brier score is known as the root Brier score (RBS). This has the property that it is a robust estimator and is an upper bound of the canonical calibration error. For this reason, it is recommended to be reported alongside bin or kernel metrics when the number of samples is small and those alternative metrics may have high variance. RBS is compared to bin, kernel, and cumulative metrics in [22]. Of the metrics tested, only RBS and the cumulative Kolmogorov-Smirnoff metric are consistent in value with respect to data size.

### 3.4 Logarithmic metrics

Negative Log Likelihood (NLL), also known as binary cross-entropy, ignorance [10], logarithmic score [21], predictive deviance [21], or logistic loss, is calculated by the equation:

$$NLL = -\frac{1}{N}\sum_{i=1}^{N}(y_i \log(c_i) + (1-y_i)\log(1-c_i)) \quad (6)$$

The NLL takes on any nonnegative value. When the NLL is small, the model is well-calibrated. Like the Brier Score, NLL is easy to compute, but conflates accuracy and calibration [25] and is not "complete" (where a perfectly calibrated classifier would produce a metric value of zero over a hypothetical infinite-sized dataset) [7]. As with the Brier score, NLL is strictly proper and can be decomposed into reliability, resolution, and uncertainty [74]. One issue with NLL is that if the confidence is zero for the correct label for any datapoint, the NLL evaluates to infinity. This reflects the fact that a good prediction system should not assign zero probability to possible events [74]. If the confidence is a small value instead of zero, the NLL can still be very large. Thus, the metric severely penalises highly unlikely predictions, which may be an indicator of lack of calibration. However, this means that single data points can have a large effect on the overall metric value, which is an undesirable property [21]. Considering that some datasets have "label noise" where the supposed ground truth labels are incorrect for a small proportion of datapoints, this property of the metric has potential to cause major issues.

The multi-class variant of NLL, also known as multinomial logistic loss, is defined as follows, where $I[y = k]$ is the indicator function, which is unity if $y = k$ and zero otherwise:

$$NLL = -\frac{1}{NK}\sum_{i=1}^{N}\sum_{k=1}^{K} I[y_i = k] \log c_{ik} \quad (7)$$

Focal loss (FL) is a modified version of NLL designed to focus on hard-to-classify examples [68]. Focal loss is defined as:

$$FL = -\frac{1}{NK}\sum_{i=1}^{N}\sum_{k=1}^{K} I[y_i = k](1-c_{ik})^\gamma \log c_{ik} \quad (8)$$

When $\gamma > 0$, FL down-weights the loss for high-confidence predictions, putting more emphasis on hard examples that are misclassified. The user-defined focussing parameter $\gamma$ is usually constant across all data points, often taking the value $\gamma = 2$. However, Mukhoti et al. [51] make it depend on the confidence value and Wang et al. [81] set it on a per-data-point basis as part of a meta-learning framework. Focal loss was originally introduced as a loss function for object detection model training and its use improves calibration. However, the loss can be used as a calibration metric in its own right. If $\gamma = 0$ then this reduces to NLL. Focal loss is usually only computed for the top-confidence class rather than the full class computation in (8). Dual focal loss is a modified version of FL that examines the top two most confident classes to take into account the extent to which the top-confidence class leads the other classes [71]. The factor $(1-c_{ik})$ is modified to be $(1-c_{ik1}+c_{ik2})$, where $k_{\{1,2\}}$ are the two most confident classes. Using dual FL as a training target results in models that are better calibrated than standard FL. Focal loss is not strictly proper, but it can be made so by adding a scaled version of the Brier score to it to make new metric, focal calibration loss (FCL) [44].

Sumler et al. [70] describe a general metric to measure the consistency of continuous or discrete probabilistic algorithms, including multi-target trackers, classifiers, multi-hypothesis trackers, and particle filters. It is shown that for the single target tracking case this is equivalent to normalised estimation error squared (NEES), a widely used metric in the tracking literature. They derive a simplified version of this metric when applied to binary classifiers and call it the entropic calibration difference (ECD). This is defined as:

$$ECD = \frac{1}{N}\sum_{i=1}^{N}(c_i - y_i)\log\left[\frac{c_i}{1-c_i}\right] \quad (9)$$

ECD is a signed metric – it can take on any value on the real line. When positive, the classifier is over-confident. When negative, the classifier is under-confident. When zero, the classifier is perfectly calibrated. The signed nature of the metric is an advantage compared to other metrics that only give information on whether or not a classifier is calibrated but not the direction of miscalibration. The absolute value of ECD is a proper scoring rule.

## 3.5 Global metrics

Global metrics compute the mean confidence of all data points and compare this to the mean accuracy. The two numbers are expected to match for a calibrated system. This type of metric is included as a point based metric due to its use of simple sums over the data points. However, it could also be considered to be a bin-based metric with a single bin for the entire dataset. Individual metrics vary in how they compare the global sums.

The global squared bias (GSB) is defined as:

$$GSB = \frac{1}{K}\sum_{k=1}^{K}\left(\frac{1}{N}\sum_{i=1}^{N} c_i - \frac{1}{N}\sum_{i=1}^{N} y_i\right)^2 \tag{10}$$

The GSB is a measure of the match between the unconditional mean predicted probability and the unconditional mean probability of the outcome [17]. The metric, as defined in (10), takes on values between zero and unity. However, when comparing GSB with the old definition of Brier score, many authors omit the $1/K$ scale factor in binary problems to keep both metrics on the scale of zero to two. Due to its global nature, the GSB provides only a crude measure of calibration error, as the metric can be zero if different parts of the reliability diagram are un-calibrated but cancel out overall. The GSB is equivalent to the expected calibration error (see section 4.2) with a single bin for all probabilities.

A metric similar to GSB is the multi-class difference of confidence and accuracy (MDCA) [28]. It is defined as:

$$MDCA = \frac{1}{K}\sum_{k=1}^{K}\left|\frac{1}{N}\sum_{i=1}^{N} c_{ik} - \frac{1}{N}\sum_{i=1}^{N} y_{ik}\right| \tag{11}$$

MDCA was designed to be used as an additional loss term for mini-batches of data in neural net training to encourage calibrated models. However, it can be used to assess calibration of the whole dataset. The metric is differentiable, which allows it to be used as part of gradient descent algorithms.

Another global metric for binary problems is the ratio of the expected to observed (EO) number of data points in the target class:

$$EO = \frac{\sum_{i=1}^{N} c_i}{\sum_{i=1}^{N} y_i} \tag{12}$$

The metric takes on non-negative values on the real line. If $EO = 1$ the model is calibrated, if $EO < 1$ the model is under-confident, and if $EO > 1$, the model is over-confident. This metric is unusual in the sense that it gives a measure of the direction of calibration as well as miscalibration in general. The few other metrics that have this property normally take on any value on the real line, with zero indicating perfect calibration. The inverse of this metric is sometimes reported, in which case it is called the observed to expected (OE) ratio [64].

## 3.6 Success rate

The success rate (SR), also known as zero-one score or misclassification loss, rewards a probabilistic prediction if the most likely class materializes [21]. In case of ties, the reward is reduced proportionally to the number of most likely classes. If # denotes the number of elements (cardinality) of a set, then the SR is:

$$SR = \frac{1}{N}\sum_{i=1}^{N}\frac{1}{\#\left\{c_{ik} = \max_k c_{ik}\right\}}\sum_{k=1}^{K} I[y_i = k]I\left[c_{ik} = \max_k c_{ik}\right] \tag{13}$$

The success rate takes on values between zero and unity. The metric is proper but not strictly proper. An asymmetric version of the metric can be defined that assigns different costs to the misclassification of different classes.

## 3.7 Normalised square metrics

The Dawid-Sebastiani score (DSS) is defined in [50] as:

$$DSS = \frac{1}{N}\sum_{i=1}^{N}\left[\left(\frac{y_i - \mu_i}{\sigma_i}\right)^2 + 2\log\sigma_i\right] \quad (14)$$

In (14), $\mu_i$ is the mean of a probabilistic prediction and $\sigma_i$ is the variance. This general formulation can apply to both continuous and discrete variables. For a binary classifier, $\mu_i = c_i$ and $\sigma_i^2 = c_i(1 - c_i)$, derived from the Bernoulli distribution. DSS takes on any nonnegative value and is a proper score, although it is not strictly proper. A multivariate version of DSS that takes on any value on the real line is discussed in [84]. This version of the metric can be used to assess multi-class problems.

The normalised squared error score (NSES) is defined in [50] as:

$$NSES = \frac{1}{N}\sum_{i=1}^{N}\left[\left(\frac{y_i - \mu_i}{\sigma_i}\right)^2\right] \quad (15)$$

NSES takes on any nonnegative value. The metric is similar to the DSS, but omits the logarithmic term. However, unlike DSS, NSES is not proper. Despite its impropriety, NSES is considered in [50] to have valuable diagnostic properties, as it can be used to distinguish over-confidence ($NSES > 1$) from under-confidence ($NSES < 1$). Indeed, NSES is often used in the tracking community as a performance measure for uncertain estimates of target locations.

## 3.8 P-norm metrics and variants

The mean absolute error (MAE) is defined as:

$$MAE = \frac{1}{N}\sum_{i=1}^{N}|y_i - c_i| \quad (16)$$

The MAE is always greater than the MSE (Brier Score), except for edge cases. The metric takes on values between zero and unity. Reference [15] categorises the MAE metric as "bad" and recommends it is not worth reporting, as it measures the expected loss of a classifier that randomly chooses a class according to its confidence, without taking into account the cost of different wrong decisions.

Wu et al. [85] define a metric called expected individual calibration error (EICE). The theoretical version of this metric is the same as MAE. However, to deal with the issue that individual data points are either correct or incorrect, and $y_i \neq c_i$ even for a calibrated inaccurate classifier, they replace $y_i$ by a jack-knife estimate of $y_i$ that can in principle be equal to $c_i$, producing a proper scoring rule. According to [85], ECE measures calibration from a global perspective, does not guarantee calibration at the individual level, and can ignore rare categories. In contrast, being able to analyze calibration at the data point level with EICE is beneficial for rare categories. However, a disadvantage of EICE is that this metric requires computation of the Hessian of the model loss with respect to the model parameters, which may be computationally infeasible for large models. If access is not granted to the model's internal structure then it is not possible to compute the metric. This limits its applicability.

The pointwise $l_p$ error (PWE) [43] is a generalised metric based on the p-norm and defined as:

$$PWE_p = \left(\frac{1}{N}\sum_{i=1}^{N}|y_i - c_i|^p\right)^{1/p} \quad (17)$$

When $p = 1$, this is the mean absolute error and is equivalent to the total variation [75][83]. When $p = 2$, this is the root mean-square error, or the square-root of the Brier score (see section 3.3). When $p = \infty$, this is known as the Chebyshev norm and is equivalent to picking the maximum absolute error for a single data point. The Chebyshev norm is unstable, because it relies on a single data point and is therefore is not recommended as a useful pointwise calibration metric. Values of $p$ other than 1, 2, or $\infty$ in the $l_p$ norm are not typically used. The pointwise $l_p$ error takes on values between zero and unity for all values of $p$.

The authors de Leeuw et al. [43] define a modified version of $PWE_1$ called the L1eps error. This is defined as:

$$L1eps = \frac{1}{N}\sum_{i=1}^{N}\sqrt{(y_i - c_i)^2 + \varepsilon} \qquad (18)$$

This is similar to a pseudo Huber loss [5] (see section 3.12 for the standard Huber loss). For small errors L1eps acts like the $l_2$ error, and for large errors it acts like the $l_1$ error. This makes it robust to outliers, while having continuous derivatives for all degrees. In contrast, the standard Huber loss has a continuous first derivative but discontinuous second derivative. The L1eps error takes on values between $\sqrt{\varepsilon}$ and $\sqrt{1 + \varepsilon}$.

The hinge loss (HL) for a binary classifier where the score $c_i$ is not necessarily a probability and can take on any real number is defined in [48] as:

$$HL = \max(0, 1 - (2y_i - 1)c_i) \qquad (19)$$

The hinge loss is normally used in conjunction with support vector machines (SVMs), where $c_i$ is a score with target values of $\pm 1$ for positive and negative classes. However, when used for general probabilistic classifiers whose confidence score is in the range [0,1], the max function is unnecessary. Therefore the mean hinge loss is simply:

$$HL = \frac{1}{N}\sum_{i=1}^{N}(1 - (2y_i - 1)c_i) \qquad (20)$$

This loss function is the same as the $l_1$ loss and therefore takes on values between zero and unity. The hinge loss is not a proper loss.

### 3.9 Spiegelhalter z statistic

The Spiegelhalter z statistic is defined by [31] as:

$$z = \frac{\sum_{i=1}^{N}(y_i - c_i)(1 - 2c_i)}{\sqrt{\sum_{i=1}^{N}(1 - 2c_i)^2 c_i(1 - c_i)}} \qquad (21)$$

This statistic, under the null hypothesis of perfect calibration, is approximately distributed according to the standard normal distribution and can take on any value on the real line. Thus it is possible to perform statistical hypothesis tests. However, in tests where an already calibrated classifier was de-calibrated, the z statistic produced p-values that were less significant than those produced in the Hosmer-Lemeshow (HL) statistic [31] (see section 4.12). This suggests the z statistic is less powerful.

### 3.10 Alpha scores

Alpha scores raise individual confidence values to the power of an exponent parameterised by $\alpha$. This allows the under- or over-weighting of high confidence values compared to lower ones.

The pseudo-spherical score (PSS) is [10]:

$$PSS = \frac{1}{N}\sum_{i=1}^{N}\sum_{k=1}^{K} I[y_i = k] \frac{c_{ik}^{\alpha-1}}{\|c_i\|_\alpha^{\alpha-1}} \quad (22)$$

This reduces to the spherical score (SS) when $\alpha = 2$ and to the ignorance as $\alpha \to 1$. The PSS takes on any nonnegative value and is a strictly proper score. The spherical score was found to perform better than the quadratic (Brier) or logarithmic (ignorance) score in a weather forecast application [21]. However, the score was found to be less useful than the logarithmic score and other measures (ranked probability score, DSS, and NSES) in assessing probabilistic predictions of daily deaths from COVID-19 [50].

The power score family of metrics for parameter $\alpha > 1$ is defined as:

$$PS = \frac{1}{N}\sum_{i=1}^{N}\sum_{k=1}^{K} I[y_i = k] \left(\sum_{l=1}^{K}(\alpha - 1)c_{il}^\alpha - \alpha c_{ik}\right) \quad (23)$$

This family of scores is strictly proper and takes on values between zero and unity [14]. Typically, $\alpha = 2$, in which case the metric is known as the proper linear score (PLS), or the quadratic score [10]. Although PLS is quadratic, the name is derived from the fact that it is a modification to a naïve linear score. For binary classification problems, PLS is equivalent to the Brier score.

### 3.11 Soft F1 score

Most of the above metrics are averages of some quantity over all data points. If a classifier always produces high confidence values but makes a small number of mistakes, this indicates a bias that will be washed out when computing these metrics, since most data points will have a low calibration error. The soft F1 (SF1) score addresses the issue of severe imbalances between mistaken and correct decisions [29]. If $\mathbf{c}$, $\mathbf{y}$, and $\mathbf{1}$ respectively denote $N$-length vectors of data point confidences, labels, and unity, the soft F1 score is defined as:

$$SF1 = \frac{2(\mathbf{1} - \mathbf{c})^T(\mathbf{1} - \mathbf{y})}{\mathbf{1}^T(2 - \mathbf{c} - \mathbf{y})} \quad (24)$$

The soft F1 score takes on values between zero and unity.

### 3.12 Point metrics for isotonic regression

The authors de Leeuw et al. [43] define a number of loss functions that can be used for isotonic regression as an alternative to the standard square error (Brier score) and other point-based metrics. These include:

- **Huber loss**. This is a piecewise metric that is quadratic for small errors and linear for large ones.
- **General least-squares**. This metric is in the form $(y - c)^T W (y - c)$ for some matrix $W$.
- **Asymmetric least-squares**. This uses different weights for negative or positive values of $(y - c)$.
- **Soft insensitive loss function** (SILF). This is zero for small errors, linear for medium errors, and quadratic for large errors.

Further analysis of these metrics is beyond the scope of this review as they are not widely used for measuring classifier calibration.

### 3.13 Consistency calibration

The metrics in this section primarily focus on reliability, and are measures of mismatch between confidence and accuracy. Tao et al. [73] propose an alternative view of calibration based on the concept of consistency. Under this view, models that are highly confident should be consistent in assigning the same top-class decision when there are minor perturbations to the data. If $x$ is the input data, $\tilde{x}$ is a perturbed version of it, $d$ is some distance measure, $\varepsilon$ is a small value, $M$ is the number of Monte Carlo runs, and $k$ is the class index, the definition of consistency for a single data point is:

$$c'_k(x_i) = \frac{1}{M} \sum_{m=1}^{M} I[y(\tilde{x}_{i,m}) = k], \qquad \text{where } d(\tilde{x}_{i,m}, x_i) < \varepsilon \tag{25}$$

A classifier is said to be consistent if the consistency matches the confidence, i.e. $c'_k(x_i) = c_k(x_i)$. A probability calibration metric for consistency is not directly defined in [73]. However, such a metric could easily be constructed by replacing $y_i$ or $I[y_i = k]$ in any of the other point-based metrics in this section with $c'_1(x_i)$ or $c'_k(x_i)$ as appropriate.

## 3.14 Ranked probability score

In some classification problems the classes are ordered in some sense, for example trying to predict the integer star rating a user gives to a product based on the text of their review. In these situations, the ranked probability score (RPS) is often used. Galdran [18] defines RPS for a finite number of classes as:

$$RPS = \frac{1}{N(K-1)} \sum_{i}^{N} \sum_{k=1}^{K-1} \left[ \sum_{j=1}^{k} (y_{ij} - c_{ij}) \right]^2 \tag{26}$$

RPS can also be used to assess situations with theoretically infinite number of classes, such as when using a negative binomial distribution to model predictions of the unknown true number of deaths in an epidemic [50]. RPS is a proper score and reduces to the Brier score for $K = 2$.

RPS increases linearly with distance of estimated class from true class, whereas a quadratic penalty may be preferred. RPS also has a hidden preference for symmetric predictions. To address these issues, Galdran [18] defines a new metric called squared absolute RPS (SARPS):

$$SARPS = \frac{1}{N(K-1)} \sum_{i}^{N} \left[ \sum_{k=1}^{K-1} \left| \sum_{j=1}^{k} (y_{ij} - c_{ij}) \right| \right]^2 \tag{27}$$

Both RPS and SARPS take on values in the range zero to unity.

# 4 Bin-based metrics

## 4.1 Introduction to bin-based metrics

Point-based metrics such as the Brier score have been used to measure calibration for several decades. However, one issue with such metrics is that it is usually impossible for a classifier to achieve a perfect score under such systems. This is because if any confidence other than zero or unity is predicted for a particular data point, there will be a difference between the confidence and the label for the correct class, which is zero or unity by definition. Thus the only way for a classifier to be assessed to have perfect calibration, is if it only ever assigns a confidence of 100% to the correct class and is always correct. This is impractical for any real dataset.

An alternative approach to measuring calibration is to group data points with the same confidence values. The proportion of times the classifier makes the correct decision for each group could be compared to the confidence of the group. Under this scheme an imperfect classifier that has less than 100% accuracy may still be perfectly calibrated, if the proportions match for all confidence values.

In practice, confidence values are on a continuous scale from zero to unity and it is unlikely that that many would take on exactly the same value. To deal with this, data points with similar, but not necessarily identical, confidence values can be grouped together in bins. The proportion of correct classification can then be compared to the mean confidence of the bin, or some other single value representative of all the data points in the bin. There are a number of ways to achieve this, and this type of metric has become popular in recent years – see [24], for example. The remainder of this section describes several bin-based metrics and some of their advantages and disadvantages. The following notation specific to binned metrics is used. The number bins is $B$, the number of data points in bin $b$ is $n_b$, and the proportion of data in bin $b$ is $p_b = n_b/N$. The mean confidence of bin $b$ is $\bar{c}_b$ and the true proportion of data labels in that bin is $\bar{y}_b$. The latter two variables may relate just to class 1 in binary problems or may be vectors for multi-class problems. Unless otherwise stated, bin-based metrics take on values between zero and unity.

## 4.2 Expected calibration error

Expected calibration error (ECE) is a commonly used metric for quantifying the miscalibration of probabilistic classifiers, where lower values indicate better calibration [24]. Several authors use the word "empirical" or "estimated" instead of "expected" to describe the metric, as the ECE is not a true expectation [69].

The binary-ECE is calculated via the formula:

$$ECE = \sum_{b=1}^{B} p_b(|\bar{y}_b - \bar{c}_b|) \tag{28}$$

The standard form of ECE uses bins of equal width (EW). Thus it is sometimes referred to as ECE-EW [67]. ECE is also known as the mean absolute calibration error [42]. ECE is easy to compute and is easy to visualise: it is the absolute area between the estimated calibration "curve" and the perfect calibration line – see Fig. 3.

However, the ECE has a number of disadvantages. First, it is trivially possible to obtain a perfect ECE by randomly estimating examples according to the label distribution [45]. For example, in a binary classification problem, if 60% of the data examples are from class 1, then assigning a confidence of 60% to all items, regardless of the input features, will produce a perfectly calibrated classifier, according to ECE, but one with poor accuracy.

Second, when using fixed bins for the input confidence score, some bins may have very few or no data points. Thus the calibration error for those bins either can't be computed or has a very high variance, which is then reflected in the overall ECE. The potential sparsity of data points for some bins is demonstrated by Guo et al. [24], who provide an example histogram of the number of samples in each bin for a ResNet image classifier. Bins with low confidence values have very few samples, and the lowest two bins have no samples at all.

Third, in multi-class problems, standard ECE is computed just for the highest probability class. It may be the case that it is important to distinguish between second, third, or lower ranking classes where the assessment from the classifier under test is being combined with other information [83].

Fourth, ECE conflates calibration and sharpness when a model is highly accurate [55]. Sharpness is the desire for models to predict with high confidence. Since models are generally overconfident, this conflation is an undesirable property.

Fifth, ECE depends on the scale of probabilities. If lots of probabilities are small (e.g. 0.001), this results in a small ECE even if the achieved accuracy is also small but is many factors times the confidence (e.g. 0.01, a factor of 10 difference) [47].

Sixth, ECE is a highly discontinuous function of classifier confidence values. This makes it a difficult metric to use in gradient based optimisation schemes [37], and a small change to a single confidence value could have a large effect on the overall ECE.

Seventh, for a model with a certain fixed calibration performance, the value of ECE decreases as the number of data points used to compute it increases [22]. This makes it hard to compare performances of models with different size datasets.

Due to these shortcomings, a number of variations on ECE have been proposed. These are described in subsequent sections. However, we first describe some general definitions for binary and multi-class classifiers that aid discussion of such metrics.

### 4.3 Definitions for binary classification with binned metrics

The $l_p$ calibration error (CE), also known as $l_p$–ECE or mean calibration error [42], is defined as:

$$CE = \left(\sum_{b=1}^{B} p_b |\bar{y}_b - \bar{c}_b|^p \right)^{1/p} \tag{29}$$

This general definition incorporates several well-known specific metrics. Kumar et al. [38] claim that the most common value is $p = 2$, and refer to this simply as the calibration error, while Hendrycks et al. [29] call it root mean-square calibration error (RMSCE). However, $l = 1$ is also commonly used – this defines the standard ECE. Furthermore, $p = \infty$, known as the Chebyshev norm, defines the maximum calibration error (MCE), discussed in section 4.5. In tests on synthetic data with known properties, hypothesis tests based on $l_2$-ECE are shown to outperform $l_1$-ECE, in terms of missed detections of miscalibration [42]. Values of $p$ other than 1, 2, or $\infty$ are not typically used.

### 4.4 Definitions for multi-class classification with binned metrics

The mean top-label confidence for a bin $b$ is given by (30), where $c_i^{max}$ is the maximum confidence over all classes for a particular data point $i$.

$$\bar{c}_b^{max} = \frac{1}{n_b} \sum_{i \in b} c_i^{max} \tag{30}$$

The mean top-label outcome for a bin $b$ is given by (31), where $y_i^{max} = 1$ if the top-label confidence is correct, or $y_i^{max} = 0$ if the top-label confidence is incorrect, for a particular data point $i$.

$$\bar{y}_b^{max} = \frac{1}{n_b} \sum_{i \in b} y_i^{max} \tag{31}$$

The top-label classification error (TCE) [38], or "confidence-ECE" [69], is then given by (29), with $\bar{y}_b$ and $\bar{c}_b$ replaced by $\bar{y}_b^{max}$ and $\bar{c}_b^{max}$.

The class-specific mean confidence for a bin $b$ and class $k$ is given by (32), where $c_{ki}$ is the class-wise confidence for a particular data point $i$ in that bin.

$$\bar{c}_{bk} = \frac{1}{n_b} \sum_{i \in b} c_{ki} \tag{32}$$

The class-specific mean outcome for a bin $b$ is given by (33), where $\delta_{k,y_i}$ is the Kronecker delta function that is unity when its arguments are the same and zero otherwise. This represents the presence of that class in a particular data point $i$ in each bin.

$$\bar{y}_{bk} = \frac{1}{n_b} \sum_{i \in b} \delta_{k,y_i} \tag{33}$$

If $w_k \in [0,1]$ denotes the importance of each class, the marginal classification error (MCE), usually with $p = 2$, [38] is:

$$MCE = \sum_{k=1}^{K} w_k \left( \sum_{b=1}^{B} p_b |\bar{y}_{bk} - \bar{c}_{bk}|^p \right)^{1/p} \tag{34}$$

If $w_k = 1/K$ for equally important classes, and $p = 1$, this is also known as class-wise ECE (CWCE) [69], macro subset ECE (MSECE) [56], or static calibration error (SCE) [55]. If $w_k$ is in proportion to the number of data points in class $k$, and $p = 1$, it is known as the weighted subset expected calibration error (WSECE) [56]. The unweighted terms of the outer sum in (34), may be considered class-stratified versions of the generalised ECE metric [66].

For a true multiclass-ECE, probability vectors could in theory be binned in simplex space. The difference between the mean probability vector and the vector of class proportions in each bin could then be computed. However, in practice, since the number of bins would be very high, most bins would likely be empty or have very few data points [69]. Thus this is not a practical measure and is not considered further here.

## 4.5 Variants of expected calibration error

The adaptive calibration error (ACE) is computed similarly to ECE. However, instead of using equal widths for the confidence bins, the bins are adaptively based on fixed percentiles of the confidence scores in the test dataset so that each bin has the same number of data points [55]. The definition of ACE is computed over all classes for multi-class problems, like the static calibration error. ACE is also called the equal-mass (EM) ECE, or ECE-EM [67]. It is shown in [67] that binning-based estimators with bins of equal mass have lower bias than estimators with bins of equal width. EW and EM methods of binning are also referred to as width binning and frequency or quantile binning, respectively [69]. Choosing a data-dependent binning scheme is known to provide much faster convergence than fixed bins [75]. Furthermore, the calibration error approximated by the use of fixed bins is a lower bound, which underestimates true miscalibration [75]. A disadvantage of ECE-EM is that some parts of the confidence space may have quite wide bins, preventing the ability to model variation in accuracy in those regions. The equal-area ECE (ECE-EA), or "equiareal ECE", has bins with approximately equal area, and this provides a middle ground between ECE-EM and ECE-EW [66].

Thresholded Adaptive Calibration Error (TACE) is computed in a similar way to ACE but only includes data points with a confidence above a certain threshold [55]. The logic behind this is that in situations with many classes, a lot of the class probabilities are low, and this washes out the calibration score. A threshold of 0.01 is used in [55]. When only the top $k$ confidence data points are selected, the method is called ECE@k [23]. As a heuristic, the number of bins is set to $\log_{10} k$ in [23], but this may result in a very small number of bins for small datasets. To avoid variability in choosing the parameter $k$, a modified version of the metric Avg-ECE@l is defined, which averages over ECE@k for uniformly spaced values of $k$ up to the integer $l$.

The inventors of the static calibration error, ACE, and TACE algorithms recommend that ACE generally be used in favour of the other metrics or standard ECE. However, if the number of classes exceeds 100, they recommend TACE. This is based on the fact that ACE and TACE are both relatively robust to label noise, where lower-rank predictions are more important [55].

The imbalanced calibration error (ICE) is defined in [23] as:

$$ICE = \frac{1}{Z} \sum_{b=1}^{B} p_b^\alpha |\bar{y}_b - \bar{c}_b| \qquad (35)$$

In (35), $Z = \sum_{b=1}^{B} p_b^\alpha$ is the normalisation constant and $\alpha = -\sum_{k=1}^{K} \gamma_k \log(\gamma_k)/\log(K)$ is a parameter between zero and unity that depends on the class size proportions $\gamma_k$. When the classes have equal sizes, $\alpha = 1$ and this is equivalent to ECE. The aim of the metric is to deal with imbalanced datasets by increasing the relative weight of bins with a small number of data points.

Region-balanced ECE (RBECE) is defined in [13] as:

$$RBECE = \frac{1}{|\Theta|} \sum_{b \in \Theta} |\bar{y}_b - \bar{c}_b| \qquad (36)$$

In (36), $\Theta$ is the set of eligible bins that each contain a minimum number of data points. Note that the metric gives equal weight to all bins rather than weighting them by sample size. The logic behind his metric is that in standard ECE some bins may have a small number data points and thus a high variance. Conversely, in equal mass ECE, the bins may all be concentrated in a particular part of the confidence space, for example if a model outputs a large number of high confidence estimates, the overall ECE-EM will be biased towards confidence in that part of the space. RBECE provides a middle ground between the two extremes. However, the metric does ignore data in sparse regions.

Label-binned calibration error (ECE-LB) uses binning to estimate true proportions of labels $\bar{y}_b$ but operates on individual samples $c_i$. ECE-LB is defined by [67] as:

$$ECE_{LB} = \left( \frac{1}{N} \sum_{b=1}^{B} \sum_{i \in b} |\bar{y}_b - c_i|^p \right)^{1/p} \qquad (37)$$

It can be shown that ECE-LB is at least as large as standard ECE.

The monotonic sweep calibration error ECE-SWEEP is a bin-based calibration metric. This metric chooses the largest number of bins for which the bin heights, as computed by standard ECE, are monotonic. When tested on data, it is found that the optimal bin count grows with sample size [67].

The contraharmonic expected calibration error (CECE) is based on the contraharmonic mean of the individual class subset ECEs [56]. If the ECE for a single class is $ECE_k$, then CECE is:

$$CECE = \sum_{k=1}^{K} ECE_k^2 / \sum_{k=1}^{K} ECE_k \qquad (38)$$

The squared terms in the numerator mean that the CECE is more biased towards the ECE of classes with high values and thus more severely "punishes" models that may have a few badly calibrated classes than the class-wise ECE metric, which may be considered a desirable property.

Maximum calibration error (MCE) is a similar metric to ECE but instead of measuring an average of the calibration errors, MCE measures the largest calibration error. This is useful when it is important for the model to be extremely well calibrated across a range of confidence values. It is calculated by the equation:

$$MCE = \max_b |\bar{y}_b - \bar{c}_b| \qquad (39)$$

MCE can lead to unintuitive results when there is wide variance in calibration between histogram bins, which is more likely to happen when the test set is small. In these situations, the metric is highly sensitive to the placement of bins [69]. The metric may be most suitable for safety-critical applications, where it is important to understand the worst-case calibration at any confidence level. This maximum statistic is also known as the $l_\infty$ or Cheyshev norm [43].

The max-variance mean-split (MVMS) binning strategy is described in [75]. This is a recursive partitioning scheme that splits confidence predictions in a potentially multi-class setting along the mean of the dimension with highest variance. An additional regularisation procedure sets a minimum number of data points per bin, which is set to 1000 in the experiments. MVMS is not compared to other binning strategies in [75] as it is used as standard for all experiments.

### 4.6  Fit-on-the-test ECE

For any estimated calibration map $\hat{y}(c)$, including bin-based maps, the fit-on-the-test (FOTT) calibration error ECE-FOTT is defined in [32] as:

$$ECE_{FOTT}^\alpha = \frac{1}{N} \sum_{i=1}^{N} |\hat{y}(c_i) - c_i|^\alpha \qquad (40)$$

Kängsepp et al. [32] define a binning scheme where instead of assuming constant probabilities based on the mean within a bin, bin probabilities may be non-continuous piecewise linear functions of the input confidence. This binning scheme defines $\hat{y}(c)$. They show that, under the FOTT paradigm, classical ECE is equivalent to the subfamily of such functions where the slope of each bin is unity. This leads to the "tilted-roof" reliability diagram, where the roofs of the bins all have angles of 45 degrees to the horizontal. This makes it easier to assess the amount of miscalibration visually in comparison to the perfect calibration identity line, also at 45 degrees. An example of a tilted-roof reliability diagram is shown in Fig. 5, for the same input data as Fig. 4.

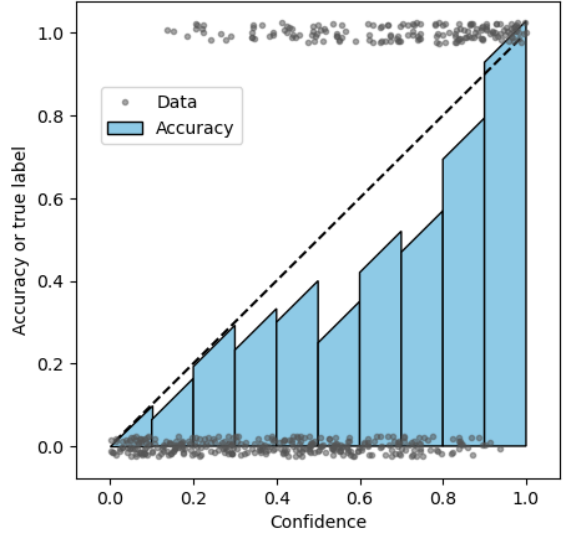

**Fig. 5** Tilted-roof reliability diagram for 500 labelled data points, with ten bins.

The number of bins to be used for ECE can be selected through cross-validation by optimising the ECE-FOTT loss. The optimum number of bins for a calibration task containing 5000 data points was 14. It is interesting that this is in the range of 10, 15, or 20 bins that are usually arbitrarily used as part of standard ECE calculations. This suggests the number of bins typically used is sensible.

### 4.7 Interval calibration error

The interval calibration error (ICE) is a theoretical metric that averages the ECE over all possible bin widths and locations. In practice, it cannot be computed directly, as the number of possibilities is huge for large datasets. However, a surrogate ICE (SICE) metric can be computed through Monte Carlo sampling in two stages [7]. The first stage computes a metric called the random interval calibration error (RICE), which is based on a modification of the bin-based ECE. For a single Monte Carlo run of RICE, all bin widths, except two, are set according to $w = 2^{-k}$ for some integer $k$. The first bin width $r$ is chosen randomly in the range $[0, w]$ and this determines the final bin width, as all bin widths must sum to unity. RICE is then defined for $B$ bins, each having an interval denoted by $Bin_b(2^{-k}, r_m)$ and containing $n_b$ data points, and $M$ Monte Carlo runs as:

$$RICE(k) = \frac{1}{M} \sum_{m=1}^{M} \sum_{b=1}^{B} \frac{1}{n_b} \sum_{i \in b} |y_i - c_i| I[c_i \in Bin_b(2^{-k}, r_m)] \qquad (41)$$

A maximum number of bins to test is chosen via $k^* = \lfloor -\log_2(\varepsilon/2) \rfloor$ for a maximum calibration error $\varepsilon = 0.01$. SICE is then computed as the optimisation over a series of exponentially smaller bin widths:

$$SICE = \min_{k=0,\ldots k^*}(RICE(k) + 2^{-k}) \quad (42)$$

The random aspect of SICE averages out the effect of discontinuous jumps in the calibration map at bin edges and thus makes it a consistent estimator, but may make it undesirable for repeatable assurance applications. In tests, SICE is better than standard ECE but not as good as the Laplace kernel calibration error (see section 5.3) or the smooth calibration error (see section 5.6).

## 4.8 Expected signed calibration error

The expected signed calibration error (ESCE) [78] or miscalibration score (MCS) [1] is:

$$ESCE = \sum_{b=1}^{B} p_b(\bar{y}_b - \bar{c}_b) \quad (43)$$

The standard definition of this metric uses equal width bins. To reduce the well-known high local variance of binned estimators, Verhaeghe et al. [78] use the mean of this metric over uniformly distributed bin sizes in the range 0.005 to 0.05. This averaging process is also used by the authors for computing standard ECE, a process with some similarities to SICE. Since ESCE can be positive or negative in the range $[-1,1]$ it can be used to assess the direction of bias, i.e. whether a classifier is over-confident (ESCE<0) or under-confident (ESCE>0). The magnitude of ESCE is equal to or less than ECE. If $|ESCE| < ECE$ this indicates the classifier is over-confident and under-confident in different parts of the confidence space.

For multi-class problems, a variant metric exists, called weighted subset MCS (WSMCS). This computes the MCS individually for each class. All under- or over-confident classes according to this metric are grouped into one of two groups. For each under- or over-confident group, a group-wise MCS is computed as a weighted sum of the class MCS values, with the weights in proportion to the class size (a micro-average). The overall WSMCS is the weighted sum of the two group-MCS values, with the weights in proportion to the number of classes in each group (a macro average). This metric takes into account the imbalance between the numbers of under- and over-confident classes. However, under- and over-confident classes can cancel each other out, so a low absolute value may mask high levels of miscalibration in individual classes.

## 4.9 Soft bin metrics

Soft-binning ECE (SBECE) uses soft binning to obtain a metric that is differentiable, allowing it to be used as a loss function to encourage a calibrated model while training the model using gradient descent [33]. The first step is to define a model to classify a confidence value as belonging to a particular bin. If $\bar{c}$ is the vector of bin centres, then the vector of bin probabilities, or bin membership function, is:

$$u(c_i) = softmax\left(\frac{-(c_i - \bar{c})^2}{\tau}\right) \quad (44)$$

In (44), $\tau$ is a temperature parameter that controls the softness of the binning. As $\tau \to 0$ the output tends to a one-hot vector, which is equivalent to standard hard binning. Bohdal et al. [33] show that SBECE is relatively insensitive to temperature parameter values $\tau$ in the range 0.0001 to 0.01.

The soft-binned size, confidence, and accuracy and of bin $b$ are:

$$s_b = \sum_{i=1}^{N} u_b(c_i) \quad (45)$$

$$c_b = \frac{1}{s_b}\sum_{i=1}^{N} u_b(c_i)c_i \quad (46)$$

$$a_b = \frac{1}{s_b} \sum_{i=1}^{N} u_b(c_i) y_i \tag{47}$$

In a similar manner to the mean calibration error in (29), the SBECE is then computed as:

$$SBECE = \left( \sum_{b=1}^{B} \frac{s_b}{N} |a_b - c_b|^p \right)^{1/p} \tag{48}$$

Differentiable ECE (DECE) is similar to SBECE, but uses a different bin membership function [9]. If the upper edges of bin $b$ are $\beta_b$, weights are defined as $w_1 = [1, 2, \ldots, B]$, and biases are defined as $w_o = \left[0, -\beta_1, -(\beta_1 + \beta_2), \ldots, -\sum_{i=1}^{B-1} \beta_b \right]$, then the probability of a confidence $c_i$ belonging to bin $b$ is:

$$u(c_i) = softmax\left( \frac{(w_1 c_i + w_0)}{\tau} \right) \tag{49}$$

Like SBECE, DECE is relatively insensitive to temperature parameter values $\tau$ in the range 0.0001 to 0.01. However, DECE better approximates ECE than SBECE and produces better calibrated models, as measured by ECE, when used during training. Both soft-bin metrics are reminiscent of kernel-based metrics (see section 5), with the membership functions over bins taking the role of kernel functions over data points. An alternative differentiable ECE metric uses the LogSumExp function to soften the choice of class with the maximum confidence value [80].

## 4.10 De-biased bin-based estimators

The so-called "plugin estimator" (PL) of the $l_2$ calibration $ECE_{PL}$ error in (29) is biased [38]. It is possible to construct a de-biased (DB) estimator by subtracting an approximate error term. For the $l_2$ error, the de-biased square estimator is:

$$CE_{DB}^2 = \sum_{b=1}^{B} p_b \left( |\bar{y}_b - \bar{c}_b|^2 - \frac{\bar{c}_b(1 - \bar{c}_b)}{n_b - 1} \right) \tag{50}$$

For the $l_1$ error, the de-biased estimator is approximated by:

$$ECE_{DB} = ECE_{PL} - \left( \left[ \frac{1}{M} \sum_{m=1}^{M} \sum_{b=1}^{B} p_b |R_{bm} - \bar{c}_b| \right] - ECE_{PL} \right) \tag{51}$$

In (51), $R_{bm} \sim N\left(\bar{y}_b, \frac{\bar{y}_b(1-\bar{y}_b)}{n_b}\right)$ is a random sample from the normal approximation of the label distribution in bin $b$, and $M$ is the number of Monte Carlo samples. The double summation computes an approximation of the expected value of the plugin ECE estimator and the large parentheses contain the overall bias.

In the limit of infinite size datasets, both metrics take on values between zero and unity. However, for finite datasets, both metrics can take on small negative values, due to the de-biasing term.

Using the plugin calibration error estimator to test for calibration leads to rejecting well-calibrated models too often [38]. That is, there are too many false alarms when attempting to detect miscalibrated models [83]. Therefore, the de-biased estimator should be used for more refined hypothesis testing.

Xiong et al. [86] show that the confidence scores for data in sparse areas of the feature space tend to be over-confident, and scores in dense areas tend to be under-confident. If the scores for dense and sparse data lie in the same confidence bin then these can cancel out making a model seem more calibrated according to ECE than it really is. To mitigate this proximity bias or cancellation effect, the proximity-informed ECE (PIECE) metric is proposed. The metric bins data points by both confidence value and "proximity" value, where proximity is based on the mean distance of a data point to its ten nearest neighbours. PIECE is defined as:

$$PIECE = \sum_{h=1}^{H} \sum_{b=1}^{B} p_{bh} |\bar{y}_{bh} - \bar{c}_{bh}| \qquad (52)$$

The bins are equal mass, with the number of confidence bins $B = 15$ and number of proximity bins $H = 10$. For the same dataset and model, PIECE is always at least as large as standard ECE, making it a stricter score. The metric can be used to assess the ability of recalibration algorithms to address the proximity bias problem. However, it has the disadvantage that it needs access to feature vectors, which may not be available in all settings. Since bins are two-dimensional, each bin has fewer data points than ECE making the variance in bins higher.

Yang et al. [87] introduce the partitioned calibration error (PCE). This operates in a similar way to PIECE, but is more general as it allows any grouping of data points based confidence or feature values, and also includes the possibility of averaging over different partitions (ways of grouping) of the same dataset. If there are $Q$ partitions of the data, each partition has $B_q$ groups, and a general loss function $L$ is defined, PCE is defined as:

$$PCE = \sum_{q=1}^{Q} p(q) \sum_{b=1}^{B_q} p_b \, L(\bar{y}_b, \bar{c}_b) \qquad (53)$$

PCE takes on many other metrics as special cases. For example, if there is one partition of the dataset into $B$ bins of confidence values, and the loss function is $L = |y - c|$, then this is standard ECE. If there is one partition of the dataset putting each data point into its own group and $L = (y - c)^2$ then this is the Brier score. Other point-, bin-, and kernel-based metrics may also similarly be obtained. For notational convenience, (53) shows the mean accuracy $\bar{y}_b$ and confidence $\bar{c}_b$, but other statistics may also be used. PCE allows the cancellation effect to be mitigated by grouping the data appropriately. However, as with PIECE, it has the disadvantage that grouping needs access to feature vectors. Issues to do with bias and loss relating to grouping effects are discussed more in [61].

## 4.11 Bias analysis

A framework known as bias by construction (BBC) is used in [67] to analyse the bias of various metrics. In this framework, the probability estimates produced by classifiers for real data are used to fit a parametric model for the density of confidence values and calibration curves. The fitted models are then used as ground truth to generate large amounts of synthetic data. The framework has been used to compare ECE, ECE-DEBIAS, ECE-SWEEP and kernel density estimation (KDE) metrics. KDE is described in section 5.4 below. The analysis also compares equal mass and equal width versions of the ECE metrics. In all cases, equal mass versions of ECE show a lower level of bias than the equal width versions as the number of samples increases. The ranking of bias, for realistic calibration curves, from least to most biased is: ECE-SWEEP, ECE-DEBIAS, ECE, KDE. Poor performance for KDE could be due to the standard parameters of the metric, which may only be optimised for the simple synthetic datasets used in [88]. For perfectly calibrated classifiers, ECE-DEBIAS is better than ECE-SWEEP. Therefore, there is little to choose between ECE-DEBIAS and ECE-SWEEP. For either method, at least 500 samples are required to reliably detect a classifier with a 10% calibration error and 10,000 samples are required to detect one with a 2% error.

Whether equal-width bins or equal data size bins are used, standard binned metrics like ECE increase their value as the number of bins increase, for a fixed number of data points. The expected values of binned metrics stay bounded away from zero if the number of data points per bin remains bounded on some fixed range of scores, as the maximum bin width becomes arbitrarily small [2].

## 4.12 Hypothesis-test bin-based metrics

Hypothesis tests can be developed to assess the significance of differences of a measured calibration metric from zero, under the assumption that a model is well calibrated. These tests are usually based on a test statistic, which can on its own be considered to be a metric for calibration. Hypothesis tests can be developed for most metrics through bootstrapping schemes. This section includes bin-based metrics that were specifically designed with hypothesis testing in mind.

The Hosmer-Lemeshow (HL) statistic has been a popular calibration measure of binary classifiers, especially in the medical field [31]. It was originally used with logistic regression classifiers [42] but has wider applicability. The "c-statistic" version is equivalent to an equal mass binning scheme with $B = 10$ bins. The "h-statistic" version is equivalent to an equal width binning scheme with the same number of bins. For either scheme, the test statistic is usually defined as:

$$HL = \sum_{b=1}^{B} \frac{(n_b \bar{y}_b - n_b \bar{c}_b)^2}{n_b \bar{c}_b} + \frac{\left(n_b(1 - \bar{y}_b) - n_b(1 - \bar{c}_b)\right)^2}{n_b(1 - \bar{c}_b)} \quad (54)$$

This can be re-written as:

$$HL = \sum_{b=1}^{B} \frac{n_b(\bar{y}_b - \bar{c}_b)^2}{\bar{c}_b(1 - \bar{c}_b)} \quad (55)$$

The HL statistic takes on any nonnegative value and follows a chi-squared distribution with $B - 2$ degrees of freedom. From this, a standard chi-squared hypothesis test can be undertaken. A version of the HL statistic for measuring the class-wise performance of multiclass problems has also been defined [69]. The HL test has satisfactory power when used with at least 400 samples [69]. However, one disadvantage of the statistic is that it is highly sensitive to small amounts of miscalibration, especially when the number of data points is large. This makes it a more difficult metric to work with when comparing classifiers that, in real situations, are unlikely to be perfectly calibrated.

Test for calibration (T-Cal) [42] is a hypothesis test based on a de-biased plugin estimator (DPE) of $l_2$-ECE.

$$DPE = \sum_{b=1}^{B} p_b \left( |\bar{y}_b - \bar{c}_b|^2 - \frac{1}{n_b^2} \sum_{i \in b} (y_i - c_i)^2 \right) \quad (56)$$

DPE is similar in form to the de-biased ECE in (50) but with the de-bias term computed in a different way. In the limit of infinite size datasets, DPE takes on values between zero and unity. However, for finite datasets, the metric can take on small negative values, due to the de-biasing term. For multi-class problems, the binning scheme used for DPE is an equal-volume partition of the simplex. However, it is noted that other binning schemes could be used. To select the optimum number of bins, the basic T-cal test requires knowledge of the smoothness of the calibration map, which is not generally known in practice. Therefore an adaptive scheme performs hypothesis tests for a range of numbers of bins and rejects the null hypothesis of calibration if any test is rejected. In tests on synthetic data with known properties, T-cal is shown to outperform Cox's method (see section 5.11) and $l_1$-ECE, in terms of missed detections of miscalibration.

Matsubara et al. [47] define a general calibration error (GCE) metric system. This has ECE and ACE as special cases. To make a general scheme specific, one has to choose a loss function, binning scheme, and norm metric. The test-based calibration error (TCE) metric is a specialisation of GCE, as described in the following, and is good for situations with imbalanced classes [47]. For TCE, the loss is the rejection percentage under a statistical hypothesis test at an alpha confidence level. TCE examines whether each model prediction in a bin can be regarded as an outlier relative to the empirical probability for that bin. The choice of binning scheme and norm for TCE is arbitrary. However, the authors choose the $l_1$ norm and a generic method for deciding on how to arrange the bins called "Near-Optimal Bins Based on PAVA-BC" (NOBB-PAVA-BC). To achieve this, the pooled adjacent violators algorithm (PAVA) used in isotonic regression is modified with block constraints (BC) to have minimum and maximum bin sizes, and combined with other steps to form the algorithm. The authors demonstrate the value of TCE by testing it with simulated data, and under-, well-, and over-calibrated models. ECE and ACE are shown to have inconsistent behaviour whereas TCE is more consistent. The disadvantage of TCE is that it is 100 times slower to compute than the other metrics. However, the total computation time for 50,000 data points is only one minute, which is reasonable for offline scenarios for testing the calibration of models. Nevertheless the complicated definition of TCE may limit its use.

### 4.13 Overlapping-bin metrics

One of the criticisms of bin-based metrics is the discontinuity at bin boundaries. The CalBin metric addresses this issue by using overlapping bins of equal mass. The first bin is defined has the first $s$ data points. The second bin has $s$ data points $i = 2, \ldots, s + 1$. This sliding bin definition is repeated to the end of the dataset and the errors are averaged over the bins. CalBin is defined as:

$$CalBin = \frac{1}{N-s} \sum_{b=1}^{N-s} \sum_{i=b}^{b+s-1} |\bar{y}_b - c_i| \qquad (57)$$

The value for $s$ is arbitrary but [6] uses $N/10$.

Another overlapping bin metric is the k-nearest neighbours (KNN) ECE, or ECE-KNN [60]. This is defined in the same way as standard ECE, except that there is one bin per data point, and the other points in each bin are the k-nearest neighbours to the point in question, in terms of confidence value. A partially manual process for selecting k is given. ECE-KNN has a lower bias than ECE-EW, ECE-SWEEP, and ECE-DEBIAS when assessing uncalibrated models.

### 4.14 Other bin-based metrics

The harmonic calibration score (HCS) is a metric that measures both calibration (via ECE) and accuracy (A) simultaneously [72]. It is defined as:

$$HCS = (1 + \beta) \frac{A(1 - ECE)}{\beta A + (1 - ECE)} \qquad (58)$$

The parameter $\beta$ controls the balance between accuracy and calibration, with larger $\beta$ emphasising calibration more. If $\beta = 1$ then this is the harmonic mean, giving equal emphasis. The aim of the metric is to have a single performance measure to compare models. It shares this property with more standard metrics like Brier score and NLL. However, where it is necessary to measure accuracy and calibration separately, this metric cannot be used.

The well-calibration ratio (WCR) is a measure of calibration for multi-class classifiers [19]. Computation of WCR requires a few steps. First the data is partitioned into $K$ groups $T_k$, each of which contains data points with whose maximum confidence is associated with one particular class $T_k = \{(y_i, \mathbf{c}_i): c_{ik} = \max_k c_{ik}\}$. It is implicitly assumed that there is a mechanism for breaking ties in the maximum confidence. The mean confidence $CF$, for $j \in \{1, \ldots, K\}$ is given by:

$$CF_{kj} = \frac{1}{|T_k|} \sum_{t=1}^{|T_k|} c_{tj} \qquad (59)$$

The mean correctness $CR$ is given by:

$$CR_{kj} = \frac{1}{|T_k|} \sum_{t=1}^{|T_k|} I[y_i = j] \qquad (60)$$

For a well-calibrated classifiers, the $CF$ and the $CR$ should converge to each other. Therefore the WCR is defined as:

$$WCR = 1 - \frac{1}{K^2} \sum_{k=1}^{K} \sum_{j=1}^{K} |CF_{kj} - CR_{kj}| \qquad (61)$$

For a perfectly calibrated classifier, $WCR = 1$, and this decreases as the calibration worsens, with a minimum value of zero. For a binary classifier, WCR is equivalent to a two-bin bin-based metric. Therefore, it does not provide as fine-grain a measure as other bin-based metrics that have more bins.

The well-calibration ratio is maximised by having all confidence values in a partition equal to the mean correctness for that partition. This rewards confidences being pushed to the centre. In contrast, the Brier score rewards probabilities near zero or unity. To overcome these drawbacks Gebel [19] defines a "calibration measure" that combines the WCR and the BS. This is defined as:

$$Cal = \sqrt{(1 - BS^{1/2}) \times WCR} \qquad (62)$$

Although [19] is a well-cited thesis due to its introduction of the Dirichlet calibration recalibration algorithm, neither the WCR nor the combined calibration measure appear to have seen use outside of [19]. Due to their lack of wide use or analysis, it is recommended these metrics are not used.

### 4.15 Discussion of other binning issues

A general problem with bin-based schemes is that the true calibration is unmeasurable with a finite number of bins. As the number of bins is increased, the measured calibration error increases [38]. It is also the case that any binned version of calibration error underestimates the true calibration error in the limit of infinite data. However, for finite datasets ECE can under or overestimate the calibration error [67].

## 5 Kernel-based and fitted-curve metrics

### 5.1 Introduction to kernels

Among the disadvantages of bin-based metrics are the arbitrary grouping of data points into bins, and discrete jumps between bins. Kernel-based metrics seek to obtain a smooth calibration curve based on a locally weighted sum of the data points. The weighting is defined by the kernel function $k$, which integrates to unity. Kernel estimates generally perform better than bin-based metrics on various criteria, such as rate of decline of mean-squared error with sample size [17].

The general kernel estimation of the smooth calibration curve $\hat{y}(c)$ is:

$$\hat{y}(c) = \frac{1}{N} \sum_{i=1}^{N} \frac{y_i}{h_i} k\left(\frac{|c - c_i|}{h_i}\right) \qquad (63)$$

The parameter $h_i$ is known as the bandwidth, or simply the width. Large widths smooth out the estimate but are unable to model sharp changes in gradient; small widths make the estimate depend more on the local data structure, but may be prone to over-fitting. It is common for the width to be the same for all data points. Three common fixed-width methods for choosing $h$ are:

- **Normal rule of thumb**. This sets $h = 1.06\sigma N^{-1/5}$, where $\sigma$ is the standard deviation of the samples [88]. This is an optimum choice if the underlying distribution is Gaussian and Gaussian kernels are used.
- **Median heuristic**. This is the median of the distance between all pairs of samples [20].
- **Cross validation**. This divides the data into cross-validation folds and computes the performance of the system on a held-out folds while also varying values of $h$. The overall performance curve for a particular value of $h$ is computed as the mean of the performance from held-out folds. The optimum of the performance curve is used to select $h$ [17].

Adaptive widths can be used to improve on fixed-width estimates, which can be poor in sparse parts of the distribution [65]. The adaptation makes the widths larger in regions with fewer data points.

Detecting miscalibration is only possible with a finite dataset when the conditional probabilities of the classes are sufficiently smooth functions of the predicted confidences, such as kernel based ones [42]. Smoothness is implied in many calibration measurement schemes but is not usually directly addressed.

Kernel-based calibration metrics are based on the differences between the estimated calibration map and the perfect calibration line (see Fig. 2). Unless otherwise stated, kernel-based metrics take on values between zero and unity.

### 5.2 Binary classifier kernel metrics based on smooth calibration curve

The mean squared calibration error (MSCE) is defined in [17] as:

$$MSCE = \frac{1}{N} \sum_{i=1}^{N} (\hat{y}(c_i) - c_i)^2 \quad (64)$$

The Gaussian kernel function is used to compute $\hat{y}(c)$ with a width of 0.08, which was optimised through cross-validation, although analysis showed that results are not sensitive to the choice of bandwidth.

Smooth ECE (SECE) is defined in [81] as:

$$SCE = \frac{1}{N} \sum_{i=1}^{N} |\hat{y}(c_i) - c_i| \quad (65)$$

The Gaussian kernel function is used to compute $\hat{y}(c)$ with a width of 0.01, which was selected via grid search. SECE should not be confused with a different kernel metric of identical name, but denoted SMECE, described as follows, or the similarly-named smooth calibration error (SCE), described in section 5.6.

Smooth ECE (SMECE) [8] is defined in a similar manner to SECE. However, it performs smoothing on the residual $y - c$ rather than the label $y$, which, in addition to a number of other design choices, gives it better mathematical properties. The metric uses a reflected Gaussian kernel to deal with edge effects in the interval [0,1], which alleviates bias from the standard Gaussian kernel. SMECE is a consistent calibration measure and is efficient with respect to both sample complexity and runtime. Crucially, it is hyperparameter-free, as a specific scheme is used to choose an optimal kernel bandwidth parameter. A grid-based computation of the metric over $G$ grid points is:

$$SMECE = \frac{1}{NG} \sum_{g=1}^{G} \left| \sum_{i=1}^{N} (y_i - c_i) k_{\sigma*}(c_i, g/G) \right| \quad (66)$$

The kernel bandwidth parameter $\sigma$ is the standard deviation of the un-reflected Gaussian kernel. Its optimum value is set so that $\sigma^* = SMECE$, which can be determined efficiently via binary search. The paper [8] describes an efficient method for computing an approximate version of SMECE based on sub-sampling the data and performing a convolution with a fast Fourier transform. Uncertainty quantification of SMECE can be determined via bootstrapping. The method behind the metric can be used to produce an associated reliability diagram, including a confidence interval for accuracy values at each input confidence value.

Popordanoska et al. [62] introduce the Dirichlet kernel density estimator (DKDE) to measure strong calibration error in multiclass problems, and call it ECE-KDE. For the binary classification problem, a partially de-biased beta kernel density estimator (BKDE) variant is defined as:

$$ECE_{BKDE} = \frac{1}{N} \sum_{j=1}^{N} \frac{\left[\sum_{i \neq j} k(c_j, c_i) y_i\right]^2 - \sum_{i \neq j} [k(c_j, c_i) y_i]^2}{\left[\sum_{i \neq j} k(c_j, c_i)\right]^2 - \sum_{i \neq j} [k(c_j, c_i)]^2} \quad (67)$$

The beta kernel is defined as:

$$k(c_j, c_i) = c_j^{\alpha_i - 1} (1 - c_j)^{\beta_i - 1} \frac{\Gamma(\alpha_i + \beta_i)}{\Gamma(\alpha_i)\Gamma(\beta_i)} \quad (68)$$

In (68), $\alpha_i = \frac{c_i}{h} + 1$ and $\beta_i = \frac{1 - c_i}{h} + 1$ and $h$ is the bandwidth parameter.

The more general multi-class DKDE, computed using the p-norm of a vector based on the label and confidence vectors, is:

$$ECE_{DKDE} = \frac{1}{N} \sum_{j=1}^{N} \left| \frac{\sum_{i \neq j} k(c_j, c_i) y_i}{\sum_{i \neq j} k(c_j, c_i)} - c_j \right|_p^p \quad (69)$$

The Dirichlet kernel is defined as:

$$k(c_j, c_i) = \frac{\Gamma(\sum_{k=1}^{K} \alpha_{ik})}{\prod_{k=1}^{K} \Gamma(\alpha_{ik})} \prod_{k=1}^{K} c_{jk}^{\alpha_{ik} - 1} \quad (70)$$

In (70), $\alpha_{ik} = \frac{c_{ik}}{h} + 1$ and $h$ is the bandwidth parameter, as before. A de-biasing scheme is available for $p = 1$ or $p = 2$.

The advantages of BKDE and DKDE are that they are consistent estimators (unlike ECE or MMCE), scalable with respect to number of classes (unlike ECE or Mix-n-Match), de-biased (unlike Mix-n-Match or MMCE), and differentiable (unlike ECE). The property of differentiability allows it to be used as a target in gradient-based training algorithms.

## 5.3 Pairwise comparison metrics

Pairwise comparison metrics compare pairs of individual point calibration errors, through the use of a kernel to weight contribution of each summation term. The metrics vary according to the kernel used, which terms to include in the summation, and whether the metric assesses class-wise calibration or strong multiclass calibration.

The maximum mean calibration error (MMCE) is a kernel-based error introduced in [37]. The motivation behind this calibration metric is to use it as a supplementary target during classifier training. The claim is that other train-time calibration methods based on entropy penalties or temperature smoothing usefully reduce aggregate calibration error but undesirably suppress legitimately confident individual predictions. MMCE can be computed from (71).

$$MMCE = \frac{1}{N^2} \sum_{i=1}^{N} \sum_{j=1}^{N} (y_i - c_i)(y_j - c_j) k(c_i, c_j) \quad (71)$$

Choice of kernel function is arbitrary but the implementation selected in [37] is the Laplacian kernel with a width of $h = 0.4$:

$$k(c_i, c_j) = \exp\left(-\frac{|c_i - c_j|}{h}\right) \quad (72)$$

A more complicated weighted version of MMCE equalises the effect of correct and incorrect examples for multi-class problems, which result in imbalanced datasets when decomposed into binary problems. This is found to improve calibration results over using the unweighted version in (71). Although (71) is quadratic in the number of data points $N$, the training time for algorithms based on MMCE is only 10% longer than other linear metrics, like NLL. MMCE is found in [37] to be a better supplementary target to be used during training time than using the baseline of using only NLL as a target of adding an entropy penalty.

Widmann et al. [83] discuss more general kernel models and suggest that MMCE can only be used for binary classification problems. However, as identified by Kumar et al. [37], the construction of the metric allows it to be applied when assessing the top-label (highest confidence) performance of multi-class classifiers.

The Laplace kernel calibration error (LKCE) can be computed using the general kernel formula in (71), using the Laplace Kernel with fixed width $h = 1$. The Laplace kernel results in a consistent calibration measure, which is not true for the widely-used Gaussian kernel. Naïve computation of (71) takes $O(N^2)$ time. However, an approximation described by Błasiok et al. [7] allows this to be computed in $O(N)$ time. Performance curves for LKCE are similar to the smooth calibration error (SCE) (see section 5.6) but not quite so smooth. The LKCE curves are beneficially less variable than ICE.

The squared kernel calibration error (SKCE) is computed in a similar manner to MMCE [83]. However, whereas MMCE applies only to binary classifiers or top-label confidence, SKCE quantifies strong multiclass calibration and hence is more generally applicable. Several versions of SKCE have been defined, all based on a pairwise error term. This term is defined as:

$$h_{ij} = (\mathbf{y}_i - \mathbf{c}_i)^T k(\mathbf{c}_i, \mathbf{c}_j)(\mathbf{y}_j - \mathbf{c}_j) \tag{73}$$

Equation (73) has a similar form to the MMCE summand in (71). However, in (73) the $\mathbf{y}_i$ and $\mathbf{c}_i$ terms are length $K$ vectors to account for full multiclass analysis and $k$ is a $K \times K$ matrix-valued kernel. The matrix-valued kernel for SKCE is chosen as the product of the scalar Laplace kernel (72) and the $K \times K$ identity matrix. The width parameter of the kernel is chosen using the median heuristic.

The biased (B) SKCE-B is defined as:

$$SKCE_b = \frac{1}{N^2} \sum_{i,j=1}^{N} h_{ij} \tag{74}$$

This is the multiclass extension of MMCE. However, the metric is biased and takes $O(N^2)$ time to compute. Nevertheless, it is a strictly proper metric [22].

The unbiased quadratic (UQ) SKCE-UQ is defined as:

$$SKCE_{uq} = \frac{2}{N(N-1)} \sum_{1 \leq i < j \leq N}^{N} h_{ij} \tag{75}$$

This metric is unbiased but still takes $O(N^2)$ time to compute. A hypothesis test exists for the SCKE-UQ metric. However, this is based on forming a bootstrap estimate that takes $O(N^3 M)$ time to compute for $M$ Monte Carlo bootstrap samples. Computation of this statistic may be prohibitive for large datasets.

The unbiased linear (UL) SKCE-UL is defined as:

$$SKCE_{ul} = \frac{1}{[N/2]} \sum_{i=1}^{[N/2]} h_{2i-1,2i} \tag{76}$$

This metric is unbiased and only takes $O(N)$ time to compute. A hypothesis test exists for the SCKE-UL metric, based on an asymptotic approximation. If $\Phi$ is the cumulative distribution of the standard normal distribution, and $\hat{\sigma}$ is the standard deviation of the terms in the summand of (76), then the p-value of the metric under the null hypothesis of perfect calibration is:

$$P(SKCE_{ul}) = 1 - \Phi\left(\frac{SKCE_{ul}}{\hat{\sigma}\sqrt{\lfloor N/2 \rfloor}}\right) \qquad (77)$$

There are two issues with SKCE-UL. The first is that the value of the metric relies on the order in which the data points are presented to the algorithm. This may be noted from equation (76), where it can be seen that only adjacent pairs of data points contribute to the sum. This means that if the same inputs are shuffled, the computed metric may be different. This is an undesirable property. The second issue is that the metric effectively assumes the data points are randomly distributed with respect to their characteristics. However, certain data processing pipelines may sort data points by confidence value or true label. If that is the case, higher weightings will be encountered than expected on average, potentially resulting in overly high metric values.

The SKCE-UL and SCKE-UQ metrics theoretically lie in the range [0, 1]. However, for nearly perfectly calibrated models with true SKCE≈0, the metric can be slightly negative for certain arrangements of data points. This is part of the normal variance associated with computing an unbiased metric. Nevertheless, this property may harm interpretability.

A comparison is made between standard ECE and the three SKCE metrics using 10,000 synthetic datasets each containing 250 samples with known ground truth and calibrated and un-calibrated models. The ECE exhibits both negative and positive bias, whereas SKCE-B is theoretically guaranteed to be biased upward. Hypothesis testing using ECE and consistency resampling (see section 2.5) is found to be unreliable and this gets worse with more classes. The asymptotic approximations for the two unbiased SKCE metrics are good for moderate numbers of classes. However, for 100 classes, SKCE-UL exhibits some mild multimodality in its distribution of values over the datasets, and for 1000 classes it is strongly bi-modal (see Figure 27 in Appendix J.2.3 of [83]). SKCE-UQ appears to have good properties for all tests performed up to 1000 classes.

In conclusion, compared to ECE and SKCE-B, SKCE-UL may be preferred as it is unbiased, quick to compute, hypothesis tests are quick to compute, and it has reasonable performance. However, the dependency of SKCE-UL on the order of data points is a major disadvantage and it performs poorly for problems with more than 100 classes. The SKCE-UQ is preferable to SKCE-UL, as it is more stable. However, this is at the expense of potentially high computation time for very large datasets, and the Monte Carlo nature of its hypothesis test, which may be undesirable for assurance applications.

## 5.4 Smoothed kernel density estimator

Zhang et al. [88] define a smoothed kernel density estimation (SKDE) based ECE estimator. This also referred to as Mix-n-Match in [62], due to its use with the Mix-n-Match recalibration method. The metric performs kernel smoothing for both the confidence estimates and the true labels. The KDE estimate of the confidences is:

$$\tilde{c}(c) = \frac{1}{N}\sum_{i=1}^{N}\frac{1}{h_i}k\left(\frac{|c - c_i|}{h_i}\right) \qquad (78)$$

Computation of the full SKDE does not scale well with the number of classes, so it is recommended to be used with the top-label or class-wise decomposition of multi-class classifiers into binary classifiers. In the binary case, SKDE can be computed based on a grid approximation to an integral, with $G$ points, as:

$$SKDE = \tilde{y}(c) = \frac{1}{G}\sum_{g=1}^{G}|y_i - \hat{y}(g/G)|_p^p \tilde{c}(g/G) \qquad (79)$$

The triweight kernel with a fixed width based on the normal rule of thumb is used, since that kernel has been recommended for problems with limited support.

Synthetic data with known ground truth is used to compare the SKDE with standard binned ECE in [88]. SKDE is shown always to outperform binning with the difference especially large for small sample sizes.

## 5.5 Reliability map

Error correcting output codes (ECOC) are an approach to combining the opinions of many different "experts", where each expert is a binary classifier trained on a sub-problem of the full multiclass problem (see section 2.3). In an ideal case, this requires each expert to produce not only probabilities but also the uncertainty in those probabilities. When combining information from the different experts, this allows the opinion of uncertain experts to have a lower weight. Least squares ECOC (LS-ECOC) is a standard method for combining probabilities but assumes all experts are equally certain among themselves and for all data points.

Kull et al. [35] propose a "reliability map" to estimate the reliability of experts for each data point, rather than their overall reliability. This instance-wise representation enables the learning of nonlinear boundaries even when using linear base models, reduces bias, and produces better models. The reliability is used to develop a modified probability combination algorithm called LS-ECOC-Reliability (LS-ECOC-R), which outperforms LS-ECOC on all synthetic and real datasets tested.

The theoretical local reliability for a confidence $c$ is based on the ratio between the variance of estimated true probability $\sigma^2(c)$ at that point and $\hat{y}(c)(1-\hat{y}(c))$, which is the variance of the Bernoulli distribution associated with the calibration map. The theoretical reliability is defined as

$$r(c) = 1 - \frac{\sigma^2(c)}{\hat{y}(c)(1-\hat{y}(c))} \qquad (80)$$

A reliability of zero means that the estimated true probability is zero or unity. A reliability of unity means that the estimated true probability is the same as the reported confidence $c$.

For practical implementation, data points are placed into clusters of size $m = 10$ that have similar values of $c$ and $\hat{y}(c)$. An unbiased estimator of the reliability map in one cluster with mean confidence $\bar{c}$ is:

$$R(\bar{c}) = 1 + \frac{1}{m-1} - \frac{\left(\sum_i^m y_i - m\hat{y}(\bar{c})\right)^2}{m(m-1)\hat{y}(\bar{c})(1-\hat{y}(\bar{c}))} \qquad (81)$$

The overall reliability map $\hat{r}(c)$ is learnt using an Epanechnikov kernel with a fixed bandwidth and local linear regression using $R(c)$ as an input. The calibration map $\hat{y}(c)$ is learnt in a similar manner. The bandwidth for calibration map learning was set to 0.01, based on optimisation from cross-validation. The bandwidth for reliability map learning was set to $m$ times this value, at 0.1. Since local linear regression can produce estimates outside the range [0, 1], and extreme values near 0 or 1 are undesired, the estimates are clipped to 0.001 or 0.999.

To test the reliability map approach, synthetic data with a Gaussian distribution was used so that the true calibration and reliability maps are known. The reliability map approach consistently outperformed a constant reliability estimator across all dataset sizes, with 400 being the smallest size tested.

The clustering approach used in reliability map estimation reduces the effective sample size by a factor of $m$ making reliability maps harder to estimate than calibration maps. Typically, 2000 data points are required for reliability map estimation when used as part of a wider processing chain, but sometimes only as few as 400 are needed.

## 5.6 Smooth calibration error

The empirical smooth calibration error (SCE) can be computed as:

$$SCE = \max_{\{z_i\}} \frac{1}{N} \sum_{i=1}^{N} (y_i - c_i) z_i \text{ subject to} \quad (82)$$
$$-1 \leq z_i \leq 1, \forall i$$
$$|z_i - z_j| \leq |c_i - c_j|, \forall i,j$$

The constraints in the definition of SCE produce an implicit weighting function $z$ that is 1-Lipschitz smooth, that is, the magnitude of the local gradient is never larger than unity. The Lipschitz condition smooths out the contribution from each neighbourhood of $c$. Unlike standard ECE, this metric is a consistent calibration measure. It provides smoother performance curves than LKCE and is less variable than ICE [7]. The SCE is unusual among other metrics in that it takes on values in the range $[-1, 1]$. The only other metric that has this property is the bin-based expected signed calibration error. This allows both of these signed metrics to measure under- or over-confidence in addition to the degree of miscalibration.

## 5.7 Logit Smoothed ECE

Logit smoothed ECE (LSECE) is designed to avoid the problems of discontinuities in ECE [11]. The metric randomly samples $M$ data points with replacement. The confidence for each sample is converted to logit space via $h(c) = \log(c/(1-c))$. A normally distributed zero-mean random variable $\sigma z_m$ with standard deviation $\sigma$ is added to the logit. A kernel estimate of the calibration curve $\hat{y}$ is formed in confidence logit space using a Gaussian kernel of width $\sigma$.

$$\hat{y}_m = \frac{\sum_{i=1}^{N} y_i k(h(c_i), h(c_m) + \sigma z_m; \sigma)}{\sum_{i=1}^{N} k(h(c_i), h(c_m) + \sigma z_m; \sigma)} \quad (83)$$

The randomised logits are mapped back to linear space by the inverse logit function $\rho$. The LSECE is then computed as the mean absolute difference between the curve and the identity function.

$$LSECE = \frac{1}{M} \sum_{m=1}^{M} |\hat{y}_m - \rho(h(c_m) + \sigma z_m)| \quad (84)$$

The randomisation process makes the LSECE metric continuous from a theoretical standpoint and in practice the metric is relatively insensitive to the arbitrary parameter $\sigma$, which is analogous to the bin width in ECE. However, the Monte Carlo nature of the metric may be undesirable for assurance applications.

## 5.8 Integrated calibration index and error percentile

The integrated calibration index (ICI) is a fitted-curve-based metric, and as such, it has some similarities with kernel-based metrics. It is similar to Cox's intercept and slope method (see section 5.11) in the sense that it fits a calibration curve and then analyses the curve [31].

The development of the ICI was motivated by Harrell's Emax index, which is the maximum absolute error (MAE) between a smooth calibration curve and the diagonal line of perfect calibration [3]. The smoothed curve is obtained via the locally estimated scatterplot smoothing (LOESS) algorithm. This is a non-parametric regression algorithm that is also known as the Savitzky-Golay filter when independent variables are a fixed width apart. The algorithm fits a low-degree polynomial to data points near to the point of interest. Austin et al. [3] utilise a two-degree polynomial, 75% of the full dataset to contribute to each estimate, and a tri-cubic weighting to down-weight data points far from the estimation point.

The ICI is the weighted difference between observed and predicted probabilities, in which observations are weighted by the empirical density function of the predicted probabilities. From a theoretical perspective, ICI is given by (85), where $\hat{y}(c)$ is the smoothed estimate of the label proportion for a particular confidence (i.e. the calibration map) and $\phi$ is the density function of the distribution of predicted probabilities.

$$ICI = \int_0^1 |c - \hat{y}(c)|\phi(c)dc \qquad (85)$$

The empirical ICI is:

$$ICI = \frac{1}{N}\sum_{i=1}^{N} |c_i - \hat{y}(c_i)| \qquad (86)$$

A metric related to ICI is the error percentile (EP) metric [3]. This is the $Pth$ percentile of the absolute difference between observed and predicted probabilities. Common values of $P$ are 90 and 50 to give the E90 and E50 metrics. E50 is the median absolute difference and E90 is the 90th percentile of this difference. Confidence intervals for ICI and EP can be estimated using bootstrap methods.

ICI, EP, and Emax have advantages over other ways of measuring calibration: they have a simple interpretation and assign a greater weight to dense data areas, which supresses poor estimates from sparse areas. The three metrics were compared by Austin et al. [3] using simulated data with known ground truth while examining the performance of correctly and incorrectly specified models. ICI tends to demonstrate more consistent behaviour in tests than EP or Emax.

## 5.9 Estimated Calibration Index

The estimated calibration index (ECI) [76], also known as the expected calibration index in [46], is similar in concept to ICI. The definition is:

$$ECI = \frac{1}{NK}\sum_{k=1}^{K}\sum_{i=1}^{N}|c_{ik} - \hat{y}(c_{ik})|^2 \qquad (87)$$

In (87), the calibration map $\hat{y}$ is estimated as a multinomial logistic regression model using a cubic spline. This equation is also reminiscent of the Brier score, where $\hat{y}$ is replaced by the actual data labels. The advantage of ECI over BS is that is measures calibration only, rather than calibration and discrimination combined.

## 5.10 Fit on the test

Kängsepp et al. [32] define a general fit-on-the-test (FOTT) paradigm, where parameters of a calibration function from a family of functions are fit to the data by minimising the ECE-FOTT loss in (40) through cross validation. Section 4.5 describes how bin-based schemes can be considered to be a particular family of functions, leading to the tilted-roof reliability diagram. Other function families have also been assessed using this paradigm.

The piecewise linear (PL) method for evaluating calibration fits a PL calibration map $\hat{y}(c)$, where parameters of the function are the number of segments, segment boundaries, and the value of the function at the boundaries. A separate calibration map is estimated for each cross-validation fold and an ensemble average is used to determine the final $\hat{y}(c)$. The ECE-PL metric for calibration is then given by (40). The optimum number of segments for a calibration task containing 5000 data points was three.

The piecewise linear in logit-logit space (PL3) method fits a continuous piecewise linear function to logit functions $h(p) = \log(p/(1-p))$. The independent variable is $h(c)$ and the dependent variable is $h(\hat{y}(c))$. When one piece (segment) is used, this is equivalent to temperature scaling, a recalibration (calibration improvement) method [24]. When two pieces are used, this is approximately equivalent to beta scaling [36]. As before, the ECE-PL3 metric for calibration is then given by (40).

Other families tested under the FOTT paradigm include ECE-EM, Platt scaling, beta scaling, isotonic regression, spline fitting, and intra-order preserving functions.

The metrics have been assessed both with synthetic data, where the true calibration map is known, and with real data, where the true calibration map can be estimated accurately when there exists magnitudes more data than used in computing the calibration metric under test. The dataset CIFAR-5m contains 5 million synthetic images created so that models trained on CIFAR-10 have similar performance and vice versa. Metrics are assessed based on three objectives: (1) the quality of the reliability diagrams; (2) the quality of the calibration error estimates; and (3) Spearman's rank correlation between the metric ranks and the (approximately) true calibration error ranks, when assessing recalibration methods. For objective (1), PL is the best metric on average, followed by PL3 and beta scaling. However, the beta scaling rank gets worse as the number of data points increases due to the small number of parameters in the model. For objective (2), the 15-bin ECE-EM or beta metrics are the best, depending on the task. PL and PL3 also perform well. For objective (3), isotonic regression is best followed by 15-bin ECE-EM. PL is better than average and outperforms PL3.

The above assessment does not provide a clear-cut ranking of calibration metrics as the ranking is different under different objectives. However, PL generally ranks well and 15-bin ECE-EM also surprisingly ranks well, especially for the important task of ranking re-calibration methods. Since ECE-EM is now considered a "classical" method and PL is relatively easy to compute, these should be considered for use generally. In passing, it is noted that the extensive experiments by Kängsepp et al. [32] required over 10,000 hours of computer time to complete. Thus, it would be costly to recreate similar bespoke experiments for new projects.

## 5.11 Cox intercept and slope

The Cox intercept and slope (CIS) method measures calibration by performing a regression of the observed outcome against the log odds of the predicted confidence [31]:

$$\hat{y} = a + b \, logit(c) \tag{88}$$

Perfect calibration has intercept $a = 0$ (measuring calibration-in-the-large) and calibration slope parameter $b = 1$. If $a > 0$ this indicates over-confidence on average, and if $a < 0$ this indicates under-confidence on average. If $b > 1$ this indicates under-confidence for high probabilities and over-confidence for low probabilities. The converse is true for $b < 1$. Both parameters take on any value on the real line.

The CIS method has some similarities with the PL3 method, with the fitted function for CIS being equivalent to the fit for PL3 with a single segment. However, ECE-PL3 computes an ECE-style measure of the area between the fitted curve and the line of perfect calibration, whereas CIS analysis looks at the values of the intercept and slope to understand calibration. Although CIS provides more information than ECE-PL3, the need to analyse more than two variables makes CIS harder to use when assessing multiple recalibration algorithms automatically.

## 5.12 Hypothesis-test kernel or curve-based metrics

The statistical beta calibration test (SBCT) fits a beta calibration curve $\hat{y}(c)$ to the calibration dataset, computes the area between this curve and the perfect calibration identity line, and uses the result to compute a p-value in a hypothesis test of whether the fitted curve is statistically significantly different from the identity [36]. The test statistic $A$ is defined as:

$$A = \int_0^1 |\hat{y}(c) - c| dc \tag{89}$$

The p-value is approximated by the quantile of $A$ within the distribution $Gamma(4.74, 1/\sqrt{91N})$. The statistical beta calibration test process has some similarities to the fit-on-the-test paradigm (see section 5.10), with $A$ being similar to ECE-FOTT. However, ECE-FOTT is computed on a sample basis, which puts more emphasis on the regions of the calibration map with more samples, whereas the integral for $A$ is computed on a uniform grid of $c$ values. The uniform grid enables the possibility of using the above gamma distribution approximation, which is good when at least 300 calibration samples are used. The disadvantage of the SBCT is that it assumes there is a parameterisation of the beta distribution that is a good fit for the data. This may not be the case for complex calibration curves with more than one inflection point.

The parabolic Wald statistic (PWS) is defined by Galbraith et al. [17]. A hypothesis test based on this statistic fits a parabola $y \approx \sum_{j=0}^{2} \hat{\theta}^{(j)} c^j$ to the data $\{(c_i, y_i)\}$ and determines whether the fitted coefficients are significantly different from those of the linear identity function $\theta_0 = (0,1,0)^T$. If $V$ is a consistent estimator of the parameter covariance matrix, the PWS is defined as:

$$PWS = (\hat{\theta} - \theta_0)^T V^{-1} (\hat{\theta} - \theta_0) \tag{90}$$

The PWS metric takes on non-negative numbers, is asymptotically distributed according to the chi-squared distribution, and standard significance tests can be constructed based on this. This statistic assumes the calibration curve may be well approximated by a parabola. This may be the case for some datasets, but is not in general true if the calibration curve has a more complex character.

The kernel calibration conditional Stein discrepancy (KCCSD) test statistic is introduced by Glaser et al. [20]. The KCCSD test is a hypothesis test for general probability models that are not necessarily normalised. Normalisation is not an issue for classifier models, as it is trivial to normalise a set of un-normalised probabilities by dividing by the sum of those probabilities. However, this can be an issue for some regression models. KCCSD is a special case of the squared kernel calibration error (SKCE) and the maximum mean discrepancy (MMD) metrics. The test is based on a bootstrap approach to estimating quantiles and determining a significance level. Computing the KCCSD statistic involves intricate nested equations and definitions, making its implementation nontrivial. Glaser et al. [20] analyse probabilistic models in general without analysing the classification case in detail. Due to the complexity of this metric, and lack of analysis for classification it is not considered further here.

Gweon et al. [26] describe a Pearson chi-squared reliability statistic based on k-nearest-neighbours in the confidence prediction space and a Bayesian approach for estimating the expected power of the reliability test for different sample sizes. The use of nearest neighbour makes this metric reminiscent of kernel methods.

# 6 Cumulative metrics

## 6.1 Empirical cumulative calibration error

Reliability diagrams are usually based on binned estimates or sometimes kernel estimates, but the selection of width parameter for either type of representation can be arbitrary. Plotting the cumulative differences between the observed and predicted values is an alternative method that avoids the setting of arbitrary parameters. Avoiding those arbitrary choices has advantages if calibration metrics are to be used to measure compliance to regulations. The cumulative difference plot (CDP) is defined as $CDP_j = 0$ for $j = 0$ and for $j = 1, \ldots, N$ as:

$$CDP_j = \frac{1}{N} \sum_{i=1}^{j} (y_i - c_i) \tag{91}$$

For a perfectly calibrated classifier, the CDP approximates a horizontal flat line at zero.

Two types of empirical cumulative calibration error (ECCE) are defined in [2]. The first is the maximum absolute deviation (MAD) of the CDP from zero. This is known as ECCE-MAD or the Kolmogorov-Smirnoff (KS) statistic.

$$ECCE_{MAD} = \max_{1 \leq j \leq N} |CDP_j| \tag{92}$$

The second error type is the range of the CDF. This is known as ECCE-R or the Kuiper statistic.

$$ECCE_R = \max_{0 \leq j \leq N} CDP_j - \min_{0 \leq j \leq N} CDP_j \quad (93)$$

The variance of $CDP_j$ is:

$$\sigma_n^2 = \frac{1}{N^2} \sum_{i=1}^{N} c_i(1 - c_i) \quad (94)$$

The statistic $ECCE_{MAD}/\sigma_n$ is equivalent in distribution to the maximum of the absolute value of standard Brownian motion over the unit interval [0, 1]. The statistic $ECCE_R/\sigma_n$ is equivalent in distribution to the range of such Brownian motion. This equivalence to Brownian motion can be used to develop a hypothesis test for each statistic and compute p-values (significance levels) for a dataset.

In [2], the ECCE metrics are assessed and compared with standard ECE using synthetic data, where it is possible to compute ground truth statistics. It is shown that the ECCE metrics can distinguish calibrated and miscalibrated classifiers as the number of samples grows large but this is not the case for ECE. Analysis of classifiers applied to real datasets with large numbers of data points produces very low-p-values for ECCE showing that such classifiers are statistically significantly miscalibrated. However the effect size is small, as seen by the un-normalised ECCE statistics.

Gupta et al. [25] define the Kolmogorov-Smirnoff calibration metric and describe it as a "binning-free calibration measure". The final form of this metric is identical to ECCE-MAD, but it is derived in a slightly different way. A generalisation of the metric also allows the assessment of whether the $r$th most likely class is correct or the top $r$ classes are correct.

The KS error is zero for perfect calibration and unity for completely un-calibrated systems. The metric can be interpreted as a percentage difference between two distributions, which aids interpretation of the metric. KS error can be shown to be a special case of kernel based measures. The relationship between KS error and the cumulative distribution function is the same as the relationship between MCE and the binned probability density function – they are both based on the maximum difference [67].

KS error is compared to RBS, ECE, and CWCE in in [22]. Only KS and RBS are consistent in value with respect to data size. KS error is used, along with ECE, KDE-ECE, MCE, and Brier Score in [25] to assess various recalibration methods on an ImageNet recognition challenge. All metrics give similar rankings for the best recalibration methods. However, since the ground truth for this challenge is not known, it is not possible to determine which metric is best.

# 7 Object detection calibration metrics

## 7.1 Introduction to object detection calibration metrics

Most work on probability calibration for deep neural nets has used a classifier paradigm for measuring calibration – little work has been done specifically for object detection. The major difference between classification and detection is that classification has a fixed number of examples to mark as correct or incorrect and every example has a confidence score, whereas object detection algorithms can in principle declare any number of detections per image, signal, or other multi-dimensional data source. False negatives (missed detections) in object detection do not have an associated confidence, so need to be treated differently to declared detections. Furthermore, defining true negatives (correctly not producing detections when there is nothing there) in object detection is difficult – how many non-objects are there in an empty image? The number of false alarms per empty image has been used as a measure of negative detection performance [41]. However, the probability calibration equivalent of this is not obvious. Nevertheless, a few authors have proposed various calibration metrics specific to object detection. Using these metrics, it has been found that DNN-based object detectors are inherently miscalibrated for both in-domain and out-of-domain predictions [59].

In many cases, classifier-based metrics could be adapted to the object detection use case by using precision instead of accuracy. For the object detection use case, $c_i$ would be the confidence with which the model detects an object, and $y_i$ would correspond to whether or not that detection has an associated object in the ground truth data. If the object is actually there then $y_i$ is 1 but if the object does not exist then $y_i$ is 0. The $y_i$ represent class labels but classifier metrics can be expanded to multi-class detection problems through the use of one-hot encoding.

One issue with simply using a classification metric for object detection is that this treats a detection as correct or not. However, in addition to class labels, detection algorithms produce the location and size of the object in the data, usually in terms of a rectangular bounding box for images. A standard way of assessing the degree of overlap between a detection and a known true object is to compute the intersection over union (IOU) of the areas of bounding boxes [58]. A detection is deemed correct if its IOU exceeds a threshold, often set at 0.5. However, this binary classification fails to capture nuances, such as the difference between a detection with an IOU just over the threshold and one closer to unity. This is not taken into account in standard classification metrics, which would treat both detections simply as correct.

Object detection versions of ECE, MCE, and other similar bin-based metrics can be defined using precision instead of accuracy in the standard definition (28). This is an approach used by several authors (e.g. [39] [52]). A drawback of object-detection ECE is the prevalence of low-confidence detections, which cluster near zero. This skews the metric, as the first bin disproportionately affects the overall value [54]. This has motivated the development of other metrics specific to the object detection use case. The remainder of this section describes a number of such metrics, all of which take on values between zero and unity.

## 7.2 Average calibration error

The average calibration error (ACE) is a bin-based metric. It measures the average absolute difference between the precision $y_b$ and mean confidence score for each equal-width bin. ACE is defined by Neumann et al. [54] as:

$$ACE = \frac{1}{B}\sum_{b=1}^{B}|y_b - c_b| \qquad (95)$$

There are two differences between ACE and standard ECE. The first is that $y_b$ represents the precision of the object detector in a particular bin rather than the accuracy. This is a necessary difference between classifiers and detectors. The second difference is that the average in (95) gives equal weight to all bins, whereas ECE weights each bin according to the number of data points in the bin. Equal bin weights avoid the issue of many low-confident detections washing out the overall metric value. In this respect, ACE is similar to region-balanced ECE. The disadvantage of ACE is that it does not take into account the accuracy of the location of the box.

## 7.3 Detection-ECE

ACE and classifier-based metrics do not take into account the localisation information produced by detectors. Küppers et al. [39] propose variants of ECE that use up to five properties of a bounding box (confidence, x-location, y-location, width, and height) and the precision to compute the metric. The equation for detection-ECE (D-ECE) computation is the same as for ECE. However, the data points are binned across potentially all parameters. Four versions of D-ECE are defined:

- **Confidence-based D-ECE**. Considers detection confidence only, and is equivalent to ECE using precision.
- **Location-based D-ECE**. Uses confidence, x-location, and y-location.
- **Shape-based D-ECE**. Uses confidence, width, and height.
- **Full D-ECE**. Uses all five parameters: confidence, x-location, y-location, width, and height.

D-ECE enables the decomposition of calibration performance by location or shape, allowing the creation of heat maps that visualize calibration across image regions or bounding box characteristics. These heat maps reveal that object detection calibration performance is generally worse at the edges of the image than in the middle. This discovery inspires a family of conditional recalibration techniques that apply a different probability calibration transform depending on the properties of each detection, such as location in the image. The D-ECE metrics are used to show that such conditional calibration provides better calibration over a wider area of the image than standard global calibration.

The disadvantage of D-ECE is that the total number of bins is very high, especially for the full version. Even when using only five bins per dimension, the total number of bins is $5^5 = 3125$. Unless a very large test dataset is used, this results in many bins with few or zero data points. To avoid the variance this introduces, Küppers et al. [39] do not include contributions from any bin with fewer than eight data points. However, this introduces bias, as it discards a significant proportion of the data.

### 7.4 Localisation-aware calibration error

Oksuz et al. [57] define a localisation-aware expected calibration error (LAECE), which uses IOU as a measure of localisation performance. LAECE is defined as:

$$LAECE = \frac{1}{K} \sum_{k=1}^{K} \sum_{b=1}^{B} \frac{n_{bk}}{N_k} |y_{bk} IOU_{bk} - c_{bk}| \tag{96}$$

In (96) $IOU_{bk}$ is the mean IOU for bin $b$ and class $k$. The metric penalises detectors that produce high-confidence but poorly-localised bounding boxes with $IOU \ll 1$. The macro average is computed over classes rather than a micro average over detections to prevent more frequent classes dominating the error. By default $B = 25$ and equal-width bins are used.

A disadvantage of LAECE is that many noisy low-confidence detections can negatively affect computation of the metric. This is an issue for top-r survival detectors that keep up to r detections per image. As a result, Oksuz et al. [57] advocate setting a confidence threshold below which detections are rejected. The optimum threshold varies by dataset and algorithm (0.3 to 0.7 in the examples given), and should be selected based on the validation set. This adds complexity to metric computation.

Kuzucu et al. [40] argue that rejecting detections below an IOU threshold loses information about reasonable detectors that assign low confidence values to such detections. They therefore propose LAECE0, which is defined in the same way as LAECE, but with an IOU threshold of zero. The authors also define a related point-based metric, called localisation-aware adaptive calibration error (LAACE0). If $V_k$ is the number of detections in class $k$, and $v$ is the index of detections within that class, then the metric is defined as:

$$LAACE_0 = \frac{1}{K} \sum_{k=1}^{K} \sum_{v=1}^{V_k} \frac{1}{V_k} |IOU_{vk} - c_{vk}| \tag{97}$$

The advantage of LAACE over LAECE and other bin-based metrics is that it has a finer grain analysis of confidence scores by avoiding putting them into bins.

For all of the metrics in this section, they are optimised by the detector producing confidence scores equal to the IOU of the correct class. If there are no detections for a particular class then the appropriate term in the summation is set to zero.

### 7.5 Estimator of L1 calibration error for a binary object detector

Popordanoska et al. [63] describe a general family of calibration metrics that takes on D-ECE and LAECE as special cases. The family uses a similarity measure $L$ that measures the match between a detected bounding box and the ground truth box. In all examples given, this is set to IOU. The family also uses a link function $\psi$ that maps the output of $L$ into the range [0,1], denoting the correctness of a detection. If $\psi$ is a threshold function then the metric is equivalent to D-ECE. If $\psi$ is the identity, then the metric is equivalent to LAECE. The authors propose their own kernel-based special case, and call it the "Estimator of L1 calibration error for a binary object detector" (L1CBOD). If there are $V$ detections across the test dataset, $b_u$ denotes the detected bounding box coordinates, and $b_u^*$ denotes the associated ground truth box coordinates, this calibration error is defined as:

$$L1CBOD = \frac{1}{V}\sum_{v=1}^{V}\left|\frac{\sum_{u\neq v}k(c_u,c_v)\psi(L(b_u,b_u^*))}{\sum_{u\neq v}k(c_u,c_v)} - c_v\right| \quad (98)$$

The kernel $k$ is the beta kernel, making this object-detection metric an extension of the classifier metric BKDE [62]. The bandwidth of the beta kernel is determined using a leave-one-out maximum likelihood procedure, but the computed value for example datasets is not stated.

L1CBOD takes on values in the range zero to unity and is consistent, asymptotically unbiased, and differentiable almost everywhere. This latter property allows it to be used as an auxiliary loss function in gradient descent model training algorithms and helps produce a well-calibrated model.

### 7.6 Metrics that deal with false negatives

Conde et al. [12] describe D-ECE as the "go-to" calibration metric for object detection. However, they point out that it does not incorporate the effect of false negative detections. Therefore they propose the expected global calibration error (EGCE) metric. This treats false negatives as if they are false positives with a confidence of unity and then computes the confidence-only D-ECE as normal, using 15 equal-width bins. Some minor modifications to the process are made. To mitigate the influence of very low-confidence detections, EGCE applies a confidence threshold of 0.1, discarding detections below this value. In some experiments, EGCE is computed as a sum over all detections rather than an average. This is because different detectors may produce a different number of detections. Any additional low-confidence incorrect detections above the threshold would make the calibration performance of a detector appear better than its counterparts, which is an undesirable feature of a metric.

The alternative quadratic global calibration (QGC) score is a point-based metric inspired by the proper Brier score for classifiers [12]. It takes into account information from false negatives (FN) as well as true positives (TP) and false positives (FP). QGC is defined as:

$$QGC = \sum_{i\in TP,FN}(c_i-1)^2 + \sum_{i\in FP}c_i^2 \quad (99)$$

As with EGCE, QGC is sometimes computed as a sum, as in (99), and sometimes normalised by $N = \#TP + \#FN + \#FP$ to make an average.

The spherical global calibration (SGC) score is a point-based metric inspired by the proper spherical score for classifiers [12]. It also takes into account false negatives, true positives, and false positives. Let:

$$r_i = \sqrt{c_i^2 + (1-c_i)^2} \quad (100)$$

SGC is then defined as:

$$SGC = N - \sum_{i\in TP}\frac{c_i}{r_i} + \sum_{i\in FP}\frac{1-c_i}{r_i} \quad (101)$$

The effect of false negatives is implicit in the $N$ term, which increases as the number of false negatives increases. As with EGCE and QGC, SGC has a normalised version.

A comparison is made by Conde et al. [12] between D-ECE, EGCE, QGC, and SGC. Since D-ECE is not affected by the number of false negatives, it is rejected as a less useful metric. QGC and SGC have very similar behaviour. Although SGC is slightly more sensitive, QGC may be preferred due to its simplicity and similarity to the widely used Brier score. A claimed advantage of EGCE is that it has asymmetric behaviour: it rewards high-confidence true positives (which are of greater interest) more than low-confidence false positives. QGC and SGC are more symmetric than EGCE. Conde et al. recommend using both EGCE and either QGC or SGC as they offer different perspectives to the problem of measuring calibration.

# 8 Visualisation methods

## 8.1 Introduction to visualisation methods

Calibration metrics are generally a single number that represents how well a classifier is calibrated. However, it is beneficial to visualise the calibration in various ways to understand in more detail which parts of the data space are calibrated or un-calibrated. This section contains a selection of methods for visualising this information.

## 8.2 Multidimensional reliability diagrams

Vaicenavicius et al. [75] describe a multiclass alternative to the standard reliability diagrams used for binary problems, called a multidimensional reliability diagram. For three-class problems, this plots probabilities on the two-dimensional simplex. An example multidimensional reliability diagram, with 25 bins, for a reasonably well-calibrated classifier is shown in Fig. 6. As a novelty for the present paper, the probability lines for each class are coloured and the probability labels are rotated to aid interpretation. In this particular example, the classifier has a slight bias towards class 1, indicated by the fact that most arrows are pointing in an upwards direction. This example is moderately easy to interpret. However, examples with a larger number of bins (e.g. see Figure 1 of [75]), or with badly-calibrated classifiers, are much harder to interpret due to the number of overlapping arrows. Furthermore, this type of visualisation is not possible for problems with more than three classes, so its utility for analysing real data is limited. However, it is useful to illustrate the concept of multiclass calibration. Silva Filho et al. [69] use a similar idea to show the calibration map for a three-class problem. However this requires three side-by-side one-verses-all simplexes to visualise the three-dimensional output probability vector.

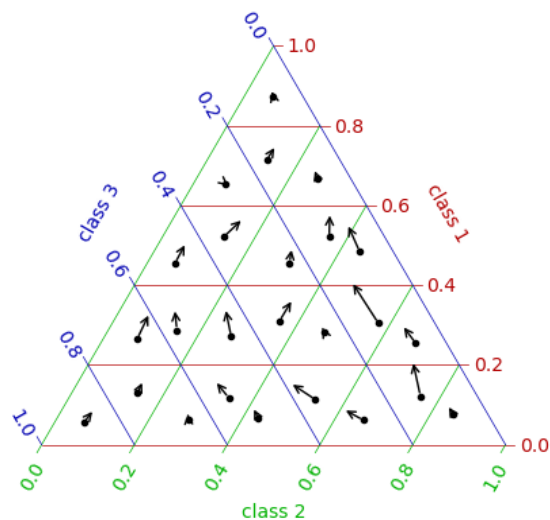

**Fig. 6** Multi-dimensional reliability diagram for a three-class classifier, with 25 bins. Arrows represent the deviation of the classifier estimated multi-class confidence (arrow head) from the true mean proportion (arrow tail) for each bin.

### 8.3 Brier curves

Hernández-Orallo et al. [30] introduce Brier curves (BC), which plot probabilistic binary classification loss as a function of cost proportion (the proportion of misclassification cost assigned to class 1 out of the total for both classes). The area under the Brier curve is the Brier score.

For binary classifiers, the probabilistic classification loss, for a particular decision threshold on confidence, is the weighted sum of the proportions of misclassifications for each class. Different costs can be assigned when misclassifying class 0 as class 1 or vice versa, which changes the weights.

The Brier curve is defined as a function of confidence $c$:

$$BC(c) = 2c\pi_0\big(1 - F_0(c)\big) + 2(1-c)\pi_1 F_1(c) \qquad (102)$$

In (102), $\pi_0$ and $\pi_1$ are the proportions of data points in classes 0 and 1, and $F_0$ and $F_1$ are the cumulative distribution functions of data points for true classes 0 or 1 being classified as class 0 (correctly or incorrectly, respectively) when the threshold is set to $c$. The two terms in (102) can be considered to be a decomposition of the BC into contribution from class 0 and class 1 respectively. An example Brier curve and its decomposition is shown in Fig. 7.

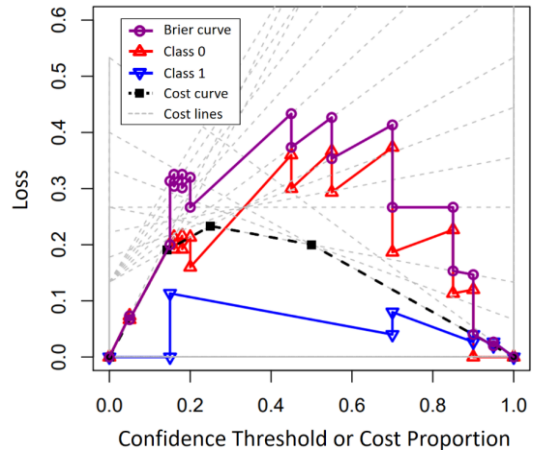

**Fig. 7** Brier curve for an example binary classifier and dataset. Adapted from [30].

An associated plot to the Bier curve is the cost curve, which shows the optimal (lowest) loss possible in the validation dataset for a given cost proportion. This would involve setting the decision threshold optimally for each data point to a value other than the confidence. This would give the minimum possible Brier score for the validation dataset. The cost curve is shown in Fig. 7 along with its associated cost lines, which each represent data-point specific decision thresholds. In practice the optimum thresholds cannot be known and the Brier score induced by the Brier curve is a more realistic measure of predicted performance.

Brier curves can be used to construct hybrid classifiers out of multiple independent classifiers by choosing the classifier with the best performance for a particular cost proportion range. This gives better performance than simply averaging the confidence values from each classifier. This construction is a non-standard ensemble method for improving calibration.

## 9 General frameworks and theory

### 9.1 Introduction to frameworks and theory

Several authors have sought to organise various calibration metrics into general frameworks or perform theoretical studies. Well-known metrics are often special cases for certain choices of specific parameters or sub-techniques within each framework. The different frameworks only partially overlap due to different forms of generalisation. No single framework encompasses all metrics. This section outlines several frameworks and how they relate to the metrics described above.

## 9.2 A unifying theory of distance from calibration

Błasiok et al. [7] describe a number of ingredients to a framework for measuring distance from calibration. These include:

- **Access model**. This refers to the data available to assess classifier calibration. Sample-access calibration measures have access to the original data $x$, as well as the predicted confidence $c$ and the true label $y$. Prediction-only access measures only have access to the confidence and label.
- **Robustness**. Robust completeness is when a calibration metric is small if a predictor is close to perfect calibration. The ECE fails this criterion as small perturbations of a perfectly calibrated predictor can result in high values of ECE. Robust soundness is when a metric is large if the predictor is badly calibrated. A consistent measure is both complete and sound. Every proper scoring rule satisfies soundness but not necessarily completeness.
- **Desiderata**. Calibration measures should be computable in the prediction-only access model, be consistent (polynomially related to the ground-truth distance from perfect calibration), and be efficient.

Błasiok et al. [7] identify three calibration error metrics that fit the above desiderata: smooth calibration (SCE), interval calibration (ICE), and Laplace kernel calibration (LKCE). These are compared to standard ECE using a synthetic dataset with a perfectly calibrated distribution that is then de-calibrated using temperature scaling of varying degrees. All metrics produce a local minimum at or near the correct temperature of unity. However, ECE shows inconsistent behaviour for high temperatures. The ICE metric, although formally consistent, has a significant undulating shape, which may limit its use. LKCE and SCE are very similar, although the SCE curve is smoother, which may lend itself better to optimisation.

## 9.3 Theoretical analysis of calibration

Bai et al. [4] provide a theoretical analysis of calibration that defines a metric "calibration error at level c". This is the true calibration map as a function of confidence and is said to allow the determination of over- or under-confidence for particular confidence values c. In general, this map is not known and must be estimated from data in some way. However, the paper uses it to analyse theoretical properties of some classifiers for which it is possible to determine the true map. Particular attention is given to logistic regression, which is shown always to be over-confident. However this is also shown to be true for any classifier with a symmetric monotonic activation function that is concave in the positive half. These findings highlight a fundamental limitation of certain widely used activation functions, suggesting that calibration issues are intrinsic to the choice of model architecture rather than purely a result of training processes or data characteristics.

## 9.4 Evaluating model calibration in classification

Vaicenavicius et al. [75] examine concept of a "calibration lens" that allows different aspects of model calibration to be examined. If a multiclass model is fully multiclass calibrated then all aspects are calibrated. However, the reverse is not necessarily true. Three lenses are examined: the canonical calibration function; the fixed partition; and the k largest predictions.

The canonical calibration function is another phrase to describe the unknown true calibration map, which is a function with estimated probabilities as inputs and true probabilities as outputs. This lens examines full multiclass calibration.

The fixed partition lens groups classes together and analyses the groups as if they are a new set of classes. This could be useful for example when grouping multiple classes into three groups whose probabilities can be analysed using a 2D multidimensional reliability diagram (see section 8.2).

The k-largest-predictions lens analyses how well a classifier is calibrated when determining whether or not the true class is in the k most likely classes as predicted by a model. This includes confidence calibration (top-label calibration) as a special case when k=1.

The calibration lens framework allows researchers to isolate and examine specific aspects of calibration, providing additional understanding that may not be apparent from aggregate calibration metrics like the standard Brier score or ECE. Each lens offers a targeted perspective, making the framework particularly useful for diagnosing calibration issues in multiclass classifiers.

# 10 Conclusion

This paper has analysed a wide range of metrics used to assess the calibration of probabilities produced by machine learning models. Table 2 in Appendix A summarizes all metrics discussed. For each metric, the table lists: alternative names; type (point, bin, kernel/curve, cumulative, or object-detection); value range (zero indicates perfect calibration unless otherwise stated); whether the metric is proper; whether a hypothesis test (HT) is available; pros and cons; and an external reference where the metric is defined or described in more detail.

The characteristics of being proper or having a hypothesis test are advantages of a metric, but are not necessary for its use. No single metric is better than all others as each one has advantages under certain conditions. Various numerical comparisons of several metrics have been carried out as described by references cited in this paper. However, each comparison only covers a small subset of all metrics described so it is difficult to assess the complete portfolio of possible metrics. This paper should facilitate making appropriate decisions on which metrics to use in different contexts.

Several open problems in probability calibration assessment are discussed in [69]. These include modelling epistemic uncertainty and assessing out-of-distribution (OOD) inputs. These topics should be considered for future research.

# 11 Declarations

**Acknowledgements:** The author thanks Emma Tattershall, Sarah Pengelly, and William Wood for productive discussions about probability calibration issues and for reviews of early versions of this survey.

**Funding**: Funding for initial literature survey and write-up activities was provided by the Defence Science Technology Laboratory under ASTRID Task 267 of contract DSTL/AGR/01142/01. Subsequent survey activities and the production of this paper were funded by the QinetiQ Fellow scheme.

**Data Availability**: Since this paper is a review article, no datasets are analysed or generated other than for illustrative purposes.

**Code Availability**: Not applicable.

**Competing interests**: The author has no competing interests to declare that are relevant to the content of this article.

**Ethical approval:** Not applicable.

**Consent to participate**: Not applicable.

**Consent for publication**: Not applicable.

# A Appendix A: Summary Table of Calibration Metrics

| Metric name | Type | Range | Proper | HT | Pros | Cons | Ref |
|---|---|---|---|---|---|---|---|
| Brier score (BS) | Point | [0,1] | Yes | No | • Widely used<br>• Easy to understand<br>• Easy to compute<br>• Robust to test set size | • Mixes calibration and accuracy<br>• Cannot achieve a perfect score without perfect accuracy | [21]<br>[69] |
| Root Brier score (RBS) | Point | [0,1] | Yes | No | • Easy to understand<br>• Easy to compute<br>• Robust to test set size | • Mixes calibration and accuracy<br>• Cannot achieve a perfect score without perfect accuracy<br>• Less used than Brier Score | [22] |
| Negative log likelihood (NLL), binary cross-entropy, ignorance, logarithmic score, predictive deviance, logistic loss | Point | $[0, \infty]$ | Yes | No | • Widely used<br>• Easy to compute | • Mixes calibration and accuracy<br>• Cannot achieve a perfect score without perfect accuracy<br>• Unstable wrt outliers or label errors | [24] |
| Focal loss (FL) | Point | $[0, \infty]$ | No | No | • Emphasises hard, misclassified examples<br>• Improves calibration more than NLL when used as a loss function. | • Adjustable focussing parameter<br>• Primarily used as a loss function rather than a metric<br>• Only computed for the top-confidence class rather than the full class computation | [68] |
| Entropic calibration difference (ECD) | Point | $[-\infty, \infty]$ | Yes | No | • Easy to compute<br>• Can assess under/over confidence (0=perfect)<br>• Theory relates to non-classifier metrics | • Dominated by points with very low or high confidence | [70] |
| Global squared bias (GSB) | Point | [0,1] | U | No | • Very easy to understand<br>• Easy to compute | • Very crude: Can be zero if different parts of the reliability diagram are un-calibrated but cancel out | [17] |
| Multi-class difference of confidence and accuracy (MDCA) | Point | [0,1] | U | No | • Very easy to understand<br>• Easy to compute<br>• Differentiable: can be used for optimisation | • Very crude: Can be zero if different parts of the reliability diagram are un-calibrated but cancel out<br>• Primarily used as a loss function rather than a metric | [28] |

| Metric name | Type | Range | Proper | HT | Pros | Cons | Ref |
|---|---|---|---|---|---|---|---|
| Expected to observed (EO) ratio | Point | $[0,\infty]$ | U | No | • Very easy to understand<br>• Easy to compute<br>• Can assess under/over confidence (1=perfect) | • Very crude: Can be zero if different parts of the reliability diagram are un-calibrated but cancel out | [64] |
| Success rate, zero-one score, mis-classification loss | Point | $[0,1]$ | Yes | No | • Very easy to understand<br>• Easy to compute | • Very crude: only check whether most likely class is correct but doesn't depend on probability | [21] |
| Dawid-Sebastiani score (DSS) | Point | $[0,\infty]$ | Yes | No | • Easy to compute<br>• Takes uncertainty in estimate into account | • Not strictly proper<br>• Approximates distribution by two moments only | [50] |
| Normalised squared error score (NSES) | Point | $[0,\infty]$ | No | No | • Easy to compute<br>• Takes uncertainty in estimate into account<br>• Can assess under/over confidence (1=perfect) | • Not widely used for classifiers | [50] |
| Mean absolute error (MAE), hinge loss | Point | $[0,1]$ | No | No | • Easy to compute | • Worse than Brier score<br>• Almost never worth reporting | [15] [34] |
| Expected individual calibration error (EICE) | Point | $[0,1]$ | Yes | No | • Enables analysis of rare classes. | • Requires access to model structure and parameters.<br>• Computationally intensive for large models.<br>• Difficult to compute. | [85] |
| Pointwise $l_p$ error | Point | $[0,1]$ | No | No | • Easy to compute | • Adjustable parameter<br>• Values other than $p \in \{1,2,\infty\}$ rarely used<br>• Chebyshev norm $p = \infty$ is not robust (depends on one data point) | [43] |
| L1eps error | Point | $\sim[0,1]$ | U | No | • Easy to compute<br>• Robust to outliers<br>• Continuous derivatives for all degrees | • Adjustable parameter<br>• Non-standard range | [43] |
| Spiegelhalter z statistic | Point | $[-\infty,\infty]$ | U | **Yes** | • Easy to compute<br>• Can assess under/over confidence (0=perfect) | • Not as powerful as HL statistic | [31] |
| Pseudo-spherical score (PSS) | Point | $[0,\infty]$ | Yes | No | • Easy to compute | • Adjustable parameter<br>• Hard to interpret meaning | [10] |

| Metric name | Type | Range | Proper | HT | Pros | Cons | Ref |
|---|---|---|---|---|---|---|---|
| Power score | Point | [0,1] | Yes | No | • Easy to compute | • Adjustable parameter<br>• More widely used PLS is a special case | [14] |
| Proper linear score (PLS) | Point | [0,1] | Yes | No | • Easy to compute | • Same as Brier score for binary classifiers | [10] |
| Soft F1 score (SF1) | Point | [0,1] | U | No | • Puts emphasis on mistakes when these are small in number | • Hard to interpret meaning | [29] |
| Ranked probability score (RPS) | Point | [0,1] | Yes | No | • Can be used for ordinal classes<br>• Can be used to assess infinite number of classes | • More widely used BS is special case for binary classes<br>• Increases linearly with true-estimated class distance<br>• Has a hidden preference for symmetric predictions<br>• Not useful for non-ordinal problems | [18] |
| Squared absolute RPS (SARPS) | Point | [0,1] | U | No | • Can be used for ordinal classes<br>• Can be used to assess infinite number of classes<br>• Increases quadratically with true-estimated class distance<br>• No preference for symmetric predictions | • Not as widely used as RPS<br>• Not useful for non-ordinal problems | [18] |
| Expected Calibration Error (ECE), estimated CE, empirical CE | Bin | [0,1] | No | No | • Easy to understand<br>• Easy to compute<br>• Widely used | • Adjustable parameter (#bins)<br>• Few data points in fixed bins: high variance.<br>• Only checks most likely class.<br>• Conflates calibration and sharpness.<br>• Depends on the scale of probabilities.<br>• Highly discontinuous function of confidence values.<br>• The most criticised metric | [24]<br>[69] |
| Adaptive calibration error (ACE), equal-mass ECE (ECE-EM) | Bin | [0,1] | No | No | • Easy to understand<br>• Easy to compute<br>• Commonly used<br>• Better than ECE | • Adjustable parameter (#bins)<br>• Conflates calibration and sharpness.<br>• Discontinuous function of confidence values | [55] |
| Thresholded Adaptive Calibration Error (TACE), ECE@k | Bin | [0,1] | No | No | • Easy to compute<br>• Better than ECE<br>• Prevents many small probabilities washing out averages | • Multiple adjustable parameters<br>• Discontinuous function of confidence values | [55]<br>[23] |

| Metric name | Type | Range | Proper | HT | Pros | Cons | Ref |
|---|---|---|---|---|---|---|---|
| Imbalanced calibration error (ICE) | Bin | [0,1] | No | No | • Easy to compute<br>• Deals with imbalanced datasets<br>• Prevents many small probabilities washing out averages | • Adjustable parameter (#bins)<br>• Discontinuous function of confidence values | [23] |
| Region-balanced ECE (RBECE) | Bin | [0,1] | No | No | • Easy to compute<br>• Deals with imbalanced datasets<br>• Prevents many small probabilities washing out averages | • Ignores some data points<br>• Multiple adjustable parameters<br>• Discontinuous function of confidence values | [13] |
| Label-binned calibration error (ECE-LB) | Bin | [0,1] | No | No | • Easy to compute<br>• Better than ECE | • Multiple adjustable parameters<br>• Discontinuous function of confidence values | [67] |
| Sweep calibration error (ECE-SWEEP) | Bin | [0,1] | No | No | • Better than ECE<br>• No adjustable parameters (unlike most bin metrics)<br>• Low levels of bias | • Discontinuous function of confidence values | [67] |
| Contraharmonic expected calibration error (CECE) | Bin | [0,1] | No | No | • Easy to compute<br>• Penalises high ECE in individual classes | • Same cons as ECE | [56] |
| Maximum calibration error (MCE) | Bin | [0,1] | No | No | • Gives worse case calibration: may be useful in safety applications | • Adjustable parameter (#bins)<br>• Few data points in fixed bins: high variance.<br>• Highly sensitive to placement of bins<br>• Highly discontinuous function of confidence values. | [24] |
| Max-variance mean-split (MVMS) | Bin | [0,1] | No | No | • Has multiclass considerations<br>• Better than ECE | • Multiple adjustable parameters<br>• Discontinuous function of confidence values | [75] |
| Fit-on-the-test ECE (ECE-FOTT) | Bin | [0,1] | No | No | • Better than ECE<br>• No adjustable parameters (unlike most bin metrics) | • Cross-validation slow and more complex to run.<br>• Discontinuous function of confidence values. | [32] |

| Metric name | Type | Range | Proper | HT | Pros | Cons | Ref |
|---|---|---|---|---|---|---|---|
| Surrogate interval calibration error (SICE) | Bin | $[0,1]$ | No | No | • Better than ECE<br>• Number of bins is optimally chosen | • Monte Carlo slow and more complex to run<br>• Monte Carlo techniques are not good for assurance<br>• Minor adjustable parameter<br>• Discontinuous function of confidence values.<br>• Not as good as Laplace kernel error | [7] |
| Expected signed calibration error (ESCE), miscalibration score (MCS) | Bin | $[-1,1]$ | No | No | • Can assess under/over confidence<br>• Averaging over number of bins makes it smoother than other bin metrics | • Discontinuous function of confidence values | [78] |
| Weighted subset miscalibration score (WSMCS) | Bin | $[-1,1]$ | No | No | • Can assess under/over confidence<br>• Can be used in multi-class problems<br>• Takes into account the imbalance between the numbers of under/over-confident classes | • Under/over-confident classes can cancel each other out, masking miscalibration in individual classes.<br>• Discontinuous function of confidence values | [78] |
| Soft-binning ECE (SBECE) | Bin | $[0,1]$ | No | No | • Differentiable: can be used for optimisation | • Adjustable parameter (temperature)<br>• Worse match to theoretical ECE than DECE | [33] |
| Differentiable ECE (DECE) | Bin | $[0,1]$ | No | No | • Differentiable: can be used for optimisation<br>• Better match to theoretical ECE than SBECE | • Adjustable parameter (temperature) | [9] |
| De-biased square calibration error (CE2-DB) | Bin | $(-\varepsilon,1)$ | U | Yes | • Unbiased metric, unlike most other bin metrics | • Unusual use of squaring compared to other bin metrics – less interpretable<br>• Same cons as ECE | [38] |
| De-biased ECE (ECE-DB) | Bin | $(-\varepsilon,1)$ | No | Yes | • Unbiased metric, unlike most other bin metrics | • Monte Carlo slow and more complex to run<br>• Monte Carlo techniques are not good for assurance<br>• Same cons as ECE | [38] |
| Proximity-informed ECE (PIECE) | Bin | $[0,1]$ | No | No | • Takes into account bias from under/over-confidences in dense/sparse regions of data space<br>• Mitigates under/over-confidence cancellation effect<br>• A stricter score than ECE | • Needs access to feature vectors, not always available<br>• Two dimensional bins mean each bin has fewer data points than ECE | [86] |

| Metric name | Type | Range | Proper | HT | Pros | Cons | Ref |
|---|---|---|---|---|---|---|---|
| Partitioned calibration error (PCE) | Bin | [0,1] | No | No | • More general than PIECE: allows any grouping of data points<br>• Allows averaging over different groupings of data<br>• Mitigates under/over-confidence cancellation effect<br>• A stricter score than ECE | • Needs access to feature vectors, not always available<br>• Multi-dimensional bins mean each bin has fewer data points than ECE or possibly PIECE | [87] |
| Hosmer-Lemeshow (HL) statistic | Bin | $[0, \infty]$ | No | Yes | • Widely used | • Adjustable parameter (#bins)<br>• Less intuitive than other bin metrics<br>• Highly sensitive to small amounts of miscalibration, for large datasets<br>• Discontinuous function of confidence values. | [31] |
| Test for calibration (T-cal) | Bin | $(-\varepsilon, 1)$ | No | Yes | • Unbiased metric, unlike most other bin metrics<br>• Use of number of bins makes it smoother than other bin metrics<br>• Better than ECE | • Complicated definition<br>• Multiple adjustable parameters<br>• Unusual use of squaring compared to other bin metrics – less interpretable<br>• Slower to compute than ECE<br>• Could be overly sensitive as null hypothesis is rejected if any number of bins rejects the hypothesis. | [42] |
| Test-based calibration error (TCE) | Bin | [0,1] | No | Yes | • Good for situations with imbalanced classes<br>• Includes method for arranging the bins<br>• Better than ECE | • Complicated definition<br>• Slower to compute than ECE<br>• Discontinuous function of confidence values | [47] |
| CalBin | Bin | [0,1] | No | No | • Averaging over overlapping bins makes it smoother than other bin metrics<br>• Better than ECE | • Adjustable parameter (#points in a bin) | [6] |
| K-nearest neighbours ECE (ECE-KNN) | Bin | [0,1] | No | No | • Averaging over overlapping bins makes it smoother than other bin metrics<br>• Lower bias than ECE-EW, ECE-SWEEP, and ECE-DEBIAS | • Requires a partially manual process for selecting k | [60] |
| Harmonic calibration score (HCS) | Bin | [0,1] | No | No | • Measures both calibration and accuracy simultaneously | • Adjustable parameter (balance between accuracy and calibration)<br>• Measures both calibration and accuracy simultaneously | [72] |

| Metric name | Type | Range | Proper | HT | Pros | Cons | Ref |
|---|---|---|---|---|---|---|---|
| Well-calibration ratio (WCR) | Bin | [0,1] | No | No | • Works with full multiclass calibration | • Complicated definition<br>• Unusual compared to other bin metrics<br>• Crude: for a binary classifier, equivalent to a two-bin bin-based metric<br>• Not mentioned outside of original reference | [19] |
| Mean squared calibration error (MSCE) | Kernel | [0,1] | No | No | • Similar definition to Brier score, a popular metric<br>• Simple definition – good interpretability<br>• Smoother than bin metrics | • Adjustable parameter (kernel width)<br>• Gaussian kernel introduces bias from edge effects<br>• Slow to compute for large datasets: $O(N^2)$ | [17] |
| Smooth ECE (SECE) | Kernel | [0,1] | No | No | • Simple definition – good interpretability<br>• Smoother than bin metrics | • Adjustable parameter (kernel width)<br>• Gaussian kernel introduces bias from edge effects<br>• Slow to compute for large datasets: $O(N^2)$<br>• Has same unabbreviated name as SMECE, which has better properties | [81] |
| Smooth ECE (SMECE) | Kernel | [0,1] | No | No | • Smoothing on residual gives better mathematical properties<br>• Reflected Gaussian kernel alleviates bias from the standard Gaussian kernel<br>• Consistent calibration measure and is efficient with respect to both sample complexity and runtime<br>• Hyperparameter-free<br>• Smoother than bin metrics<br>• Fast to compute for large datasets: $O(N)$ | • Reflected Gaussian kernel makes computation harder<br>• Slightly more complicated definition than SECE reduces interpretability<br>• Has same unabbreviated name as SECE | [8] |
| Dirichlet kernel density estimator (DKDE) | Kernel | [0,1] | No | No | • Full multiclass calibration<br>• Consistent and de-biased<br>• Differentiable: can be used for optimisation<br>• Smoother than bin metrics | • Slightly complex definition<br>• Slow to compute for large datasets: $O(N^2)$ | [62] |
| Beta kernel density estimator (BKDE) | Kernel | [0,1] | No | No | • Consistent and de-biased<br>• Differentiable: can be used for optimisation<br>• Smoother than bin metrics | • Binary or class-wise calibration only<br>• Slow to compute for large datasets: $O(N^2)$ | [62] |

| Metric name | Type | Range | Proper | HT | Pros | Cons | Ref |
|---|---|---|---|---|---|---|---|
| Maximum mean calibration error (MMCE) | Kernel | [0,1] | No | No | • Has a multiclass version<br>• Better properties than NLL – more stable<br>• Smoother than bin metrics | • Multiclass version has complex definition<br>• Biased and inconsistent metric<br>• Slow to compute for large datasets: $O(N^2)$ | [37] |
| Laplace kernel calibration error (LKCE) | Kernel | [0,1] | No | No | • Laplace kernel is consistent whereas Gaussian is not<br>• Calibration map is less up-and-down than SICE<br>• Smoother than bin metrics | • Slow to compute for large datasets: $O(N^2)$ [a linear approximation exists]<br>• Not as smooth as smooth calibration error (SCE) | [37] [7] |
| Unbiased linear squared kernel calibration error (SKCE-UL) | Kernel | $[-\varepsilon, 1]$ | No | Yes | • Unbiased<br>• Works with full multiclass calibration<br>• Fast to compute for large datasets: $O(N)$<br>• Smoother than bin metrics | • Performance degrades with number of classes starting between 100 and 1000 classes | [83] |
| Unbiased quadratic Squared kernel calibration error (SKCE-UQ) | Kernel | $[-\varepsilon, 1]$ | No | Yes | • Unbiased<br>• Works with full multiclass calibration<br>• Works well with any number of classes<br>• Smoother than bin metrics | • Monte Carlo slow and more complex to run<br>• Monte Carlo techniques are not good for assurance<br>• Slow to compute for large datasets: $O(N^2)$ | [83] |
| Smoothed kernel density estimator (SKDE) | Kernel | [0,1] | No | No | • Better than all bin metrics at assessing calibration errors<br>• Smoother than other kernel metrics and all bin metrics<br>• Fast to compute for large datasets: $O(N)$ | • Slightly complex definition<br>• Computation scales badly with number of classes, recommended as a top-label or class-wise metric<br>• Biased metric | [88] |
| Least squares error correcting output codes reliability (LS-ECOC-R) | Kernel | [0,1] | No | No | • Low bias<br>• Produces better results than some other metrics<br>• Smoother than bin metrics | • Multiple adjustable parameters<br>• Complex definition<br>• Needs more data points than other kernel methods | [35] |
| Smooth calibration error (SCE) | Kernel | [−1,1] | No | No | • Consistent calibration measure<br>• Smoother performance curves than LKCE<br>• Less up-and-down than SICE<br>• Smoother than bin metrics | • Non-trivial definition: hard to interpret meaning<br>• Slow to compute for large datasets: $O(N^2)$ | [7] |
| Logit smoothed ECE (LSECE) | Kernel | [0,1] | No | No | • Smoother than bin metrics | • Adjustable parameter (sigma)<br>• Monte Carlo techniques are not good for assurance<br>• Slow to compute for large datasets: $O(N^2)$ | [11] |

| Metric name | Type | Range | Proper | HT | Pros | Cons | Ref |
|---|---|---|---|---|---|---|---|
| Kernel calibration conditional Stein discrepancy (KCCSD) | Kernel | [0,1] | No | Yes | • Works with full multiclass calibration<br>• Smoother than bin metrics<br>• Part of a wider theory on probabilistic algorithms | • Monte Carlo slow and more complex to run<br>• Analysis specific to classifiers is not available | [20] |
| Integrated calibration index (ICI) | Curve | [0,1] | No | No | • Relatively simple definition<br>• Similar definition to ECE – helps gain acceptability<br>• Easy to compute<br>• Assign a greater weight to dense data areas than sparse ones – more stable<br>• Better than the similar EP | • Multiple adjustable parameters<br>• Slow to compute for large datasets: $O(N^2)$ | [31][3] |
| Error percentile (EP) | Curve | [0,1] | No | No | • Relatively simple definition<br>• Easy to compute<br>• Assign a greater weight to dense data areas than sparse ones – more stable | • Multiple adjustable parameters<br>• Not as good as the similar ICI<br>• Slow to compute for large datasets: $O(N^2)$ | [3] |
| Estimated calibration index (ECI), expected calibration index | Curve | [0,1] | No | No | • Similar definition to BS – helps gain acceptability<br>• Unlike BS, purely measures calibration<br>• Smoother than bin metrics | • Non-trivial definition | [76] |
| Piecewise linear ECE (ECE-PL) | Curve | [0,1] | No | No | • Better than all metrics for reliability diagram quality<br>• Better than average and outperforms PL3 when ranking recalibration methods<br>• Similar definition to ECE – helps gain acceptability<br>• No adjustable parameters<br>• Easier to explain than PL3<br>• Fast to compute for large datasets: $O(N)$ | • Cross-validation: slow to compute and slightly complex to implement | [32] |
| Piecewise linear in logit-logit space ECE (ECE-PL3) | Curve | [0,1] | No | No | • Has the popular temperature scaling as a special case<br>• Better than all metrics except ECE-PL for reliability diagrams quality<br>• Similar definition to ECE – helps gain acceptability<br>• No adjustable parameters<br>• Fast to compute for large datasets: $O(N)$ | • Cross-validation: slow to compute and slightly complex to implement<br>• Harder to explain than ECE-PL | [32] |

| Metric name | Type | Range | Proper | HT | Pros | Cons | Ref |
|---|---|---|---|---|---|---|---|
| FOTT beta scaling | Curve | [0,1] | No | No | • Better than all metrics except ECE-PL and ECE-PL3 for reliability diagram quality<br>• Better than all metrics for quality of the calibration error estimates<br>• No adjustable parameters<br>• Fast to compute for large datasets: $O(N)$ | • Cross-validation: slow to compute and slightly complex to implement<br>• Rank of performance gets worse as the number of data points increases | [32] |
| Cox intercept and slope (CIS) | Curve | $[-\infty, \infty]$<br>$[-\infty, \infty]$ | No | No | • Can assess under/over confidence<br>• Can assess more than one aspect of calibration<br>• No adjustable parameters<br>• Classic method<br>• Fast to compute for large datasets: $O(N)$ | • Multiple variables makes it hard to assess multiple recalibration algorithms automatically.<br>• Not as flexible as the more general ECE-PL3 | [31] |
| Statistical beta calibration test (SBCT) | Curve | [0,1] | No | No | • Uniform emphasis across confidence values rather than data samples<br>• Relatively easy to compute<br>• Similar definition to ECE – helps gain acceptability<br>• Fast to compute for large datasets: $O(N)$ | • The metric is only as good as the beta fit to the data | [36] |
| Parabolic Wald statistic | Curve | $[0, \infty]$ | No | Yes | • Relatively easy to compute<br>• No adjustable parameters<br>• Fast to compute for large datasets: $O(N)$ | • The metric lacks interpretability | [17] |
| Empirical cumulative calibration error maximum absolute deviation (ECCE-MAD), Kolmogorov-Smirnoff (KS) | Cumulative | [0,1] | No | Yes | • Assumption free<br>• No adjustable parameters<br>• Better than ECE<br>• Generalisation available for top r classes form of multiclass problems<br>• Easily interpretable | • Can (correctly) produce highly significant p-values for large datasets even when effect is small | [2] |
| Empirical cumulative calibration error range (ECCE-R), Kuiper | Cumulative | [0,1] | No | Yes | • Assumption free<br>• No adjustable parameters<br>• Better than ECE | • Can (correctly) produce highly significant p-values for large datasets even when effect is small | [2] |

| Metric name | Type | Range | Proper | HT | Pros | Cons | Ref |
|---|---|---|---|---|---|---|---|
| Average calibration error (ACE) | Object / bin | [0,1] | No | No | • Equal weight to all bins avoids the issue of many low-confident detections washing out metric<br>• Same pros as ECE | • Equal weight to all bins misses ability to give more weight to high-mass high-confidence bins<br>• Doesn't deal with false negatives<br>• Same cons as ECE | [54] |
| Detection-ECE (D-ECE) | Object / bin | [0,1] | No | No | • Can take into account bounding box information<br>• Allows computation of ECE heat maps vs. location<br>• Same pros as ECE | • If using bounding box information, can lead to sparse bins: high variance or missed out data<br>• Doesn't deal with false negatives<br>• Same cons as ECE | [39] |
| Localisation-aware calibration error (LAECE) | Object / bin | [0,1] | No | No | • Takes IOU bounding box information into account<br>• Takes a balanced view over all classes<br>• Same pros as ECE | • Complexity of choosing the correct IOU threshold (unless LAECE0 is used)<br>• Doesn't deal with false negatives<br>• Same cons as ECE | [40]<br>[57] |
| Localisation-aware adaptive calibration error (LAACE) | Object / point | [0,1] | No | No | • Takes IOU bounding box information into account<br>• No adjustable parameters, unlike LAECE<br>• Finer grain analysis of confidence scores than bin metrics | • Does not take balanced view over all classes, unlike LAECE<br>• Doesn't deal with false negatives | [40] |
| Estimator of L1 calibration error for a binary object detector" (L1CBOD) | Object / kernel | [0,1] | No | No | • An extension to object detection of BKDE<br>• consistent, asymptotically unbiased<br>• Differentiable: can be used for optimisation | • Adjustable parameter (kernel bandwidth)<br>• Assesses class-wise calibration only, not full calibration<br>• Doesn't deal with false negatives | [63] |
| Expected global calibration error (EGCE) | Object / bin | [0,1] | No | No | • Deals with false negatives<br>• Asymmetric behaviour: rewards high-confidence true positives more than low-confidence false positives<br>• Same pros as ECE | • Multiple adjustable parameters<br>• There is a sum and a mean version<br>• The mean version has some undesirable properties<br>• Same cons as ECE | [12] |
| Quadratic global calibration (QGC) | Object / point | [0,1] | Yes | No | • Deals with false negatives<br>• Similar to widely used Brier score<br>• Simple definition | • There is a sum and a mean version<br>• The mean version has some undesirable properties | [12] |
| Spherical global calibration (SGC) | Object / point | [0,1] | Yes | No | • Deals with false negatives<br>• SGC is slightly more sensitive than QGC | • There is a sum and a mean version<br>• The mean version has some undesirable properties | [12] |

*Table 2 Summary of calibration metrics*